\documentclass{article} % For LaTeX2e
\usepackage{iclr2022_conference,times}

\usepackage{amsmath,amsfonts,bm}

\def\eqref#1{equation~\ref{#1}}
\def\1{\bm{1}}

\DeclareMathAlphabet{\mathsfit}{\encodingdefault}{\sfdefault}{m}{sl}
\SetMathAlphabet{\mathsfit}{bold}{\encodingdefault}{\sfdefault}{bx}{n}

\usepackage{hyperref}
\usepackage{url}

\usepackage[pdftex]{graphicx}
\usepackage[utf8]{inputenc} % allow utf-8 input
\usepackage[T1]{fontenc}    % use 8-bit T1 fonts
\usepackage{hyperref}       % hyperlinks
\usepackage{url}            % simple URL typesetting
\usepackage{booktabs}       % professional-quality tables
\usepackage{amsfonts}       % blackboard math symbols
\usepackage{nicefrac}       % compact symbols for 1/2, etc.
\usepackage{microtype}      % microtypography
\usepackage{xcolor}         % colors
\usepackage{soul}
\usepackage{tabularx}
\usepackage{xspace}
\usepackage{multirow}
\usepackage{enumitem}
\usepackage[frozencache=true,cachedir=minted-cache]{minted}

\usepackage{float}
\usepackage{setspace}
\usepackage{array}

\newcommand*{\eg}{e.g.\@\xspace}

\newcommand{\mypar}[1]{\noindent{\bf #1}}
\renewcommand{\cite}{\citep}

\def\wmod{WOMD\xspace}    % Waymo Open Motion Dataset 
\def\modelname{Scene Transformer\xspace}

\title{\modelname: A unified architecture for predicting multiple agent trajectories}
\author{\textbf{Jiquan Ngiam}$^{*, 1}$, \textbf{Benjamin Caine}$^{*, 1}$,
\textbf{Vijay Vasudevan}$^{*, 1}$, \\
\textbf{Zhengdong Zhang}$^{1}$, \textbf{Hao-Tien Lewis Chiang}$^{2}$, \textbf{Jeffrey Ling}$^{2}$, \\
\textbf{Rebecca Roelofs}$^{1}$, \textbf{Alex Bewley}$^{1}$, \textbf{Chenxi Liu}$^{2}$, \textbf{Ashish Venugopal}$^{2}$, \\ \textbf{David Weiss}$^{2}$, \textbf{Ben Sapp}$^{2}$, \textbf{Zhifeng Chen}$^{1}$, \textbf{Jonathon Shlens}$^{1}$\\
$^{1}$Google Research, Brain Team,$^{2}$ Waymo
\\ {\tt \{jngiam,bencaine,vrv\}@google.com}}

\iclrfinalcopy % Uncomment for camera-ready version, but NOT for submission.
\begin{document}

\maketitle

\begin{abstract}
Predicting the motion of multiple agents is necessary for planning in dynamic environments.  This task is challenging for autonomous driving since agents (e.g., vehicles and pedestrians) and their associated behaviors may be diverse and influence one another.  Most prior work have focused on predicting independent futures for each agent based on all past motion, and planning against these independent predictions.  However, planning against independent predictions can make it challenging to represent the future interaction possibilities between different agents, leading to sub-optimal planning.
%This is a particularly acute problem for autonomous vehicles since agents (e.g., vehicles and pedestrians) and their associated behaviors may be diverse, and the decisions of the autonomous vehicle itself may influence the other agents significantly. Prior works have approached this challenge by decomposing the problem into first independently predicting futures for each agent, and subsequently planning against these fixed predictions. These approaches often suffer from an inability to represent the interaction possibilities between the different agents, leading to suboptimal planning.
% old: In this work, we formulate a model for predicting the behavior of all agents jointly in real-world driving environments. 
In this work, we formulate a model for predicting the behavior of all agents jointly, producing consistent futures that account for interactions between agents.
Inspired by recent language modeling approaches, we use a masking strategy as the query to our model, enabling one to invoke a single model to predict agent behavior in many ways, such as potentially conditioned on the goal or full future trajectory of the autonomous vehicle or the behavior of other agents in the environment.
Our model architecture employs 
% fuses heterogeneous world state in a unified Transformer architecture by employing 
attention to combine features across road elements, agent interactions, and time steps.  
We evaluate our approach on autonomous driving datasets for both marginal and joint motion prediction, and achieve state of the art performance across two popular datasets. 
Through combining a scene-centric approach, agent permutation equivariant model, and a sequence masking strategy, we show that our model can unify a variety of motion prediction tasks from joint motion predictions to conditioned prediction.

% old: Our work demonstrates that formulating the problem of motion prediction in a unified architecture with a masking strategy may allow us to have a single model that can perform motion prediction and conditioned prediction effectively.

% to rich, diverse and reusable representations, and furthermore may spur the development of new methods for jointly predicting the behavior of agents in complex environments such as autonomous vehicles.
\end{abstract}

% We seek inspiration to this problem from recent advances in zero-shot and few-shot language modeling which suggests that pretraining models on large corpora and querying the resulting representations leads to state-of-the-art performance across a range of tasks. 
\section{Introduction}

% Restore for arxiv.org.
%\blfootnote{$^{*}$ Denotes equal contribution. Please see author contributions section for details.}
Motion planning in a dense real-world urban environment is a mission-critical problem for deploying autonomous driving technology. Autonomous driving is traditionally considered too difficult for a single end-to-end learned system~\cite{thrun2006stanley}. Thus, researchers have opted to split the task into sequential sub-tasks~\cite{zeng2019end}: (i) perception, (ii) motion prediction, and (iii) planning. Perception is the task of detecting and tracking objects in the scene from sensors such as LiDARs and cameras. Motion prediction
% ~\cite{SocialGAN,alahi2016social,biktairov2020prank,buhet2020plop,casas2020spagnn,chai2019multipath,cui2019multimodal,gao2020vectornet,hong2019rules,lee2017desire,liang2020laneGCN,marchetti2020mantra,mercat2020multi,zhao2020tnt} 
involves predicting the futures actions of other agents in the scene. Finally, planning
% ~\cite{bansal2019chauffeurnet,zeng2019end,sadat2020ppp} 
involves creating a motion plan that navigates through dynamic environments. 

Dividing the larger problem into sub-tasks achieves optimal performance when each sub-task is truly independent. However, such a strategy breaks down when the assumption of independence does not hold. For instance, the sub-tasks of motion prediction and planning are not truly independent---the autonomous vehicle's actions may significantly impact the behaviors of other agents. Similarly, the behaviors of other agents may dramatically change what is a good plan. The goal of this work is to take a step in the direction of unifying motion prediction and planning by developing a model that can exploit varying forms of conditioning information, such as the AV's goal, and produce joint consistent predictions about the future for all agents simultaneously.

% ============

% In this work, we introduce an approach that performs motion prediction for multiple agents simultaneously, and can also produce predictions conditioned on the goal of the autonomous vehicle (AV).

% The goal of this work is to directly model the dependencies in order to unify a variety of motion prediction tasks from joint motion predictions to conditioned prediction.

% The goal of this work is to directly model the dependencies in order to build a unified motion prediction system that may flexibly adapt to and exploit arbitrary conditioning information -- whether for modeling real world examples, or hypothetical counterfactual scenarios.

% The goal of this work is to take a step in the direction of unifying motion prediction and planning by developing a model that can exploit conditioning information, such as the AV's goal, and produce joint consistent predictions about the future for all agents simultaneously.

% ============

% A prerequisite of such a multi-task system is that it needs to be able to jointly predict the futures of multiple agents (including the autonomous vehicle), while simultaneously taking into account their interactions. 
While the motion prediction task has traditionally been formulated around per-agent independent predictions, recent datasets~\cite{ettinger2021large,interactiondataset} have introduced interaction prediction tasks that enable us to study joint future prediction (Figure \ref{figure:motivation}). These interaction prediction tasks require models to predict the joint futures of multiple agents: models are expected to produce future predictions for all agents such that the agents futures are consistent \footnote{Marginal agent predictions may conflict with each other (have overlaps), while consistent joint predictions should have predictions where agents respect each other's behaviors (avoid overlaps) within the same future.} with one another. 

A naive approach to producing joint futures is to consider the exponential number of combinations of \textit{marginal} agent predictions. Many of the combinations are not consistent, especially when agents have overlapping trajectories. We  present a unified model that naturally captures the interactions between agents, and can be trained as a joint model to produce scene-level consistent predictions across all agents (Figure \ref{figure:motivation}, right).
% predicts a distribution over joint futures for all agents across time in a single inference pass, capturing the interactions between agents.
Our model uses a scene-centric representation for all agents ~\cite{lee2017desire,hong2019rules,casas2020spagnn,salzmann2020trajectron++} to allow scaling to large numbers of agents in dense environments. We employ a simple variant of self-attention \cite{vaswani2017attention} in which the attention mechanism is efficiently factorized across the agent-time axes. The resulting architecture simply alternates attention between dimensions representing time and agents across the scene, resulting in a computationally-efficient, uniform, and scalable architecture.

We find that the resulting model, termed {\it \modelname}, achieves superior performance on both independent (marginal) and interactive (joint) prediction benchmarks. 

Moreover, we demonstrate a novel formulation of the task using a masked sequence model, inspired by recent advances in language modeling~\cite{brown2020language, devlin2018bert}, to allow conditioning of the model on the autonomous vehicle (AV) goal state or full trajectory. In this reformulation, a single model can naturally perform tasks such as motion prediction, conditional motion prediction, and goal-conditioned prediction simply by changing which data is visible at inference time.
%We further show how we can use a masking strategy, inspired by recent advances in language modeling~\cite{brown2020language, devlin2018bert}, to train a multi-task model that performs both motion prediction, conditional motion prediction, and goal-conditioned prediction. The masking strategies allow us to adapt what data the model has access to at input time, such that the masks corresponds to different tasks.

We hope that our unified architecture and flexible problem formulation opens up new research directions for further combining motion prediction and planning. In summary, our key contributions in this work are:
%We hope that this work provides a compelling example for how predicting joint, consistent futures of all agents in an environment opens new opportunities for critical tasks in autonomous vehicle technology. In summary, our key contributions in this work are:
\vspace{-6pt}

\begin{itemize}[leftmargin=0.5cm,rightmargin=0.5cm,itemsep=2pt,parsep=2pt]

\item A novel, scene-centric approach that allows us to gracefully switch training the model to produce either marginal (independent) and joint agent predictions in a single feed-forward pass. Our model achieves state-of-the-art on both marginal and joint prediction tasks on both the Argoverse and the Waymo Open Motion Dataset.

\item A permutation equivariant Transformer-based architecture factored over agents, time, and road graph elements that exploits the inherent symmetries of the problem.  The resulting architecture is efficient and integrates the world state in a unified way.

\item A masked sequence modeling approach that enables us to condition on hypothetical agent futures at inference time, enabling conditional motion prediction or goal conditioned prediction.

% \item Through our sequence masking strategy, we show that our model can be elegantly adapted to a variety of motion prediction tasks from joint motion predictions to conditioned prediction.

% A masked sequence modeling approach that enables goal-conditioned predictions with our model.

% \item A masked sequence modeling approach that enables us to arbitrarily condition on hypothetical agent futures at inference time, enabling conditional motion prediction or goal conditioned prediction.

% \item A unified multi-task training and inference framework in which a single model can accomplish the tasks of motion prediction, conditional motion prediction, and goal-directed planning.

% Redo attempt:
% \item A novel, scene-centric approach that allows us to 
\end{itemize}

\section{Related Work}

\begin{figure*}[t]
\centering
\begin{tabular}{m{3.2cm}ccccm{3.5cm}}
     \textit{marginal prediction} &&&&& \textit{\,\,\,\,\,\,\,\,\,\,\,\,\,\,\,\,joint prediction} \\
\end{tabular}
\\
\vspace{0.03cm}
\fbox{\includegraphics[width=0.38\textwidth]{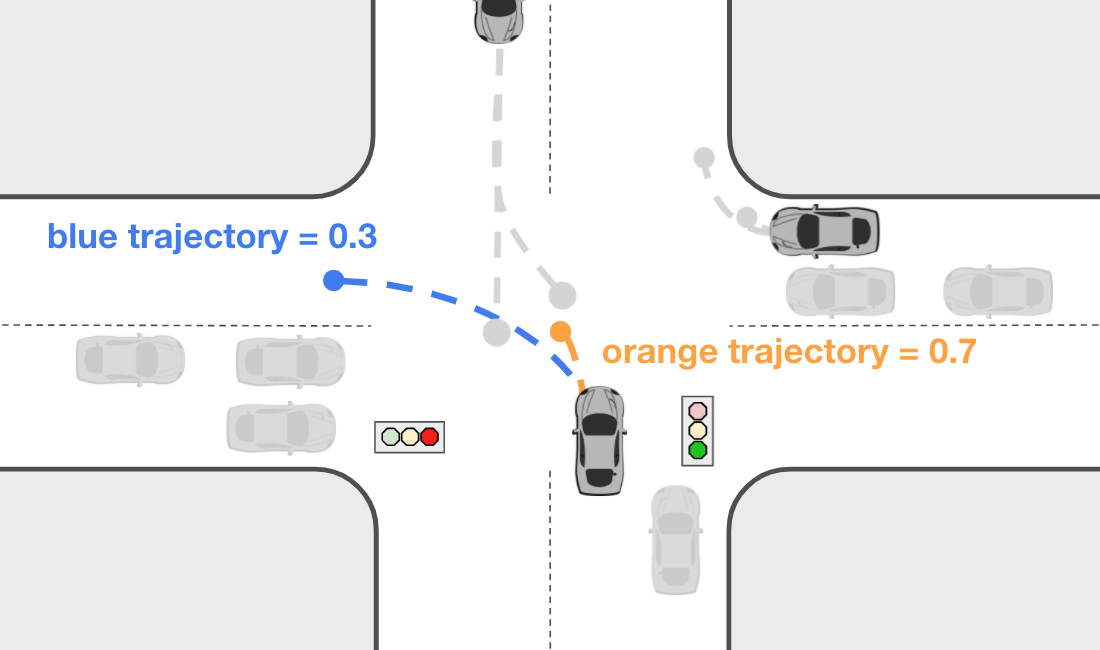}}
\hspace{0.25cm}
\fbox{\includegraphics[width=0.38\textwidth]{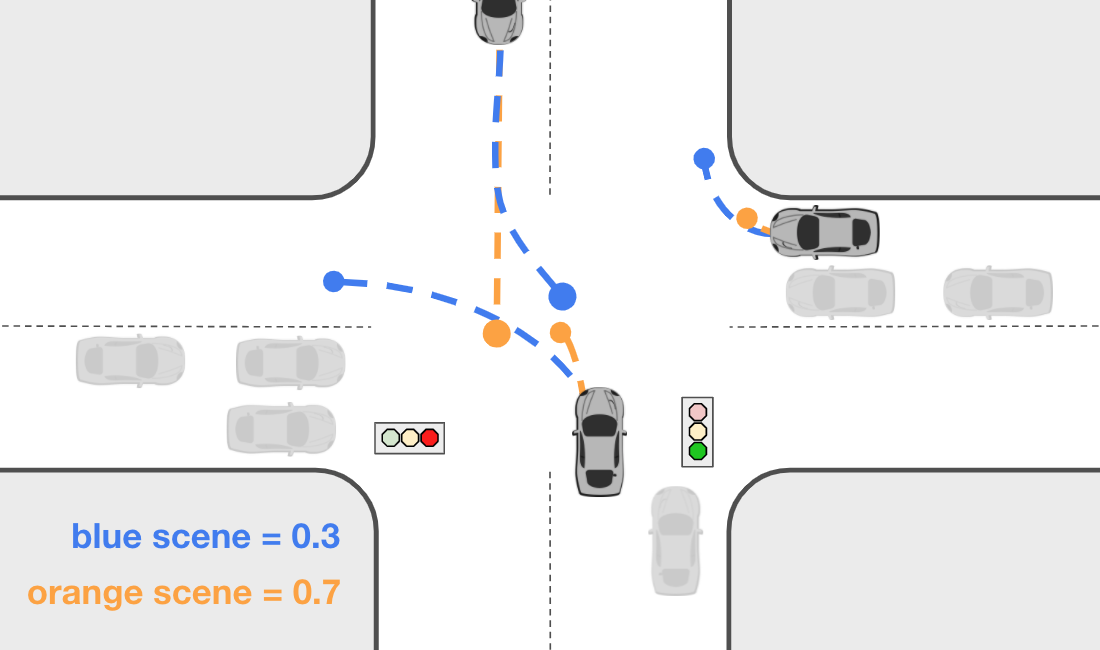}}
\caption{\textbf{Joint prediction provides consistent motion prediction.} Illustration of differences between marginal and joint motion prediction. Each color represents a distinct prediction. Left: Marginal prediction for bottom center vehicle. Scores indicate likelihood of trajectory. Note that the prediction is independent of other vehicle trajectories. Right: Joint prediction for three vehicles of interest. Scores indicate likelihood of entire scene consisting of trajectories of all three vehicles. \label{figure:motivation}}
\vspace{-0.42cm}
\end{figure*}

% We focus our discussion here on our key contribution areas in the domain of behavior prediction and goal-directed planning in a real-world driving setting.  We relate choices of architecture, frames of reference, and output representation as they relate to \modelname.

\mypar{Motion prediction architectures.} Motion prediction models have flourished in recent years, due to the rise in interest in self-driving applications and the release of related datasets and benchmarks~\cite{REF:lyftperception2019, chang2019argoverse, caesar2020nuscenes,ettinger2021large}.
%This is a particularly interesting modeling problem due to the rich and heterogeneous world state in this environment. 
Successful models must take into account the history of agent motion,  and the elements of the road graph (\eg, lanes, stop lines, traffic light dynamic state). Furthermore, such models must learn the relationships between these agents in the context of the road graph environment.
%the relationship between these agents as well as road graph elements (\eg, lanes, stop lines, traffic light dynamic state) and their relationship to the agents.
%, the history of agent motion, relationships between agents, and the road network.

One class of models draws heavily upon the computer vision literature, rendering inputs as a multi-channel rasterized top-down image~\cite{cui2019multimodal, chai2019multipath,lee2017desire,hong2019rules,casas2020spagnn,salzmann2020trajectron++, zhao2019matf}.  In this approach, relationships between scene elements are captured via convolutional deep architectures. However, the localized structure of the receptive field makes capturing spatially-distant interactions challenging.  
A popular alternative is to use an entity-centric approach.  With this approach, agent state history is typically encoded via sequence modeling techniques like RNNs~\cite{mercat2020multi, wimp2020, lee2017desire, alahi2016social, rhinehart2019precog} or temporal convolutions~\cite{liang2020laneGCN}. Road elements are approximated with basic primitives (\eg piecewise-linear segments) which encode pose information and semantic type.  Modeling relationships between entities is often presented as an information aggregation process, and models employ pooling~\cite{zhao2020tnt, gao2020vectornet, lee2017desire, alahi2016social,SocialGAN},  soft-attention~\cite{mercat2020multi, zhao2020tnt, salzmann2020trajectron++} as well as graph neural networks~\cite{casas2020spagnn,liang2020laneGCN,wimp2020}.

Like our proposed method, several recent models use Transformers~\cite{vaswani2017attention}, composed of multi-headed attention layers. Transformers are a popular state-of-the-art choice for sequence modeling in natural language processing \cite{brown2020language,devlin2018bert}, and have recently shown promise in core computer vision tasks such as detection~\cite{bello2019attention,carion2020detr,srinivas2021bottleneck}, tracking~\cite{hung2020soda} and classification~\cite{ramachandran2018stand,vaswani2021scaling,dosovitskiy2020image_transformers,bello2021lambdanetworks,bello2019attention}.  For motion modeling, recent work has employed variations of self-attention and Transformers for modeling different axes: temporal trajectory encoding and decoding~\cite{yu2020stformer,giuliari2020transformer,yuan2021agentformer}, encoding relationships between agents~\cite{li2020interactionformer, park2020diverse, yuan2021agentformer, yu2020stformer, mercat2020multi, bhat2020trajformer}, and encoding relationships with road elements.  When applying self-attention over multiple axes, past work used independent self-attention for each axis~\cite{yu2020stformer}, or flattened two axes together into one joint self-attention layer~\cite{yuan2021agentformer} -- by comparison, our method proposes axis-factored attention to model relationships between time steps, agents, and road graph elements in a unified way.

% liu2021mmformer

\mypar{Scene-centric versus agent-centric representations.}
Another key design choice is the frame of reference in which the representation is encoded. Some models do a majority of modeling in a global, scene-level coordinate frame, such as work that employs a rasterized top-down image~\cite{cui2019multimodal, chai2019multipath,lee2017desire,hong2019rules,casas2020spagnn,salzmann2020trajectron++}.  This can lead to a more efficient model due to a single shared representation of world state in a common coordinate frame, but comes with the potential sacrifice of pose-invariance.
%but making the model pose-invariant can be challenging.  
On the other hand, models that reason in the agent-coordinate frame \cite{mercat2020multi,zhao2020tnt, wimp2020} are intrinsically pose-invariant, but scale linearly with the number of agents, or quadratically with the number of pairwise interactions between agents.  Many works employ a mix of a top-down raster representation for road representation fused with a per-agent representations~\cite{rhinehart2019precog, tang2019multiple, lee2017desire}.  Similar to our own work, LaneGCN~\cite{liang2020laneGCN} is agent-centric yet representations are in a global frame -- to the best of our knowledge, this is the only other work to do so.  This enables efficient reasoning while capturing arbitrarily distant interactions and high-fidelity state representations without rasterization.
%avenugopal: perhaps its worth stating explicetly here that the frame is now shifted
% to a single agent-frame, that of the agent for whom we are planning.

\mypar{Representing multi-agent futures.}
A common way to represent agent futures is via a weighted set of trajectories per agent
\cite{ alahi2016social,  biktairov2020prank,  buhet2020plop, casas2020spagnn,  casas2020spagnn,  chai2019multipath,  cui2019multimodal,   gao2020vectornet,  hong2019rules,  lee2017desire,   marchetti2020mantra,  mercat2020multi,   salzmann2020trajectron++, zhao2020tnt, chandra2018:traphic}. This representation is encouraged by benchmarks which primarily focus on per-agent distance error metrics~\cite{caesar2020nuscenes, chang2019argoverse, interactiondataset}.  We argue in this work that modeling {\em joint} futures in a multi-agent environment (Figure \ref{figure:motivation}, right) is an important concept that has been minimally explored in prior work.  Some prior work consider a factorized pairwise joint distribution, where a subset of agent futures are conditioned on other agents -- informally, modeling $P(X)$ and $P(Y|X)$ for agents $X$ and $Y$~\cite{wimp2020, kate_cbp, salzmann2020trajectron++}.  To generalize joint prediction to arbitrary multi-agent settings, other  work~\cite{tang2019multiple,rhinehart2019precog,casas2020implicit, suo2021trafficsim,yeh2019diverse} iteratively roll out samples per-agent, where each agent is conditioned on previously sampled trajectory steps. In contrast, our model directly decodes a set of \textit{k} distinct joint futures with associated likelihoods.

\section{Methods}

\begin{figure*}[t]
\centering
\includegraphics[width=0.50\textwidth]{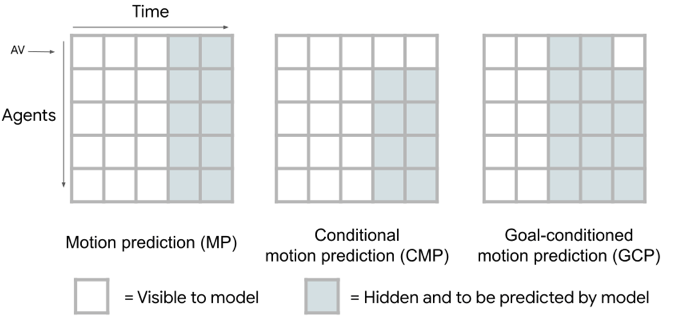} 
\hfill
\includegraphics[width=0.45\textwidth]{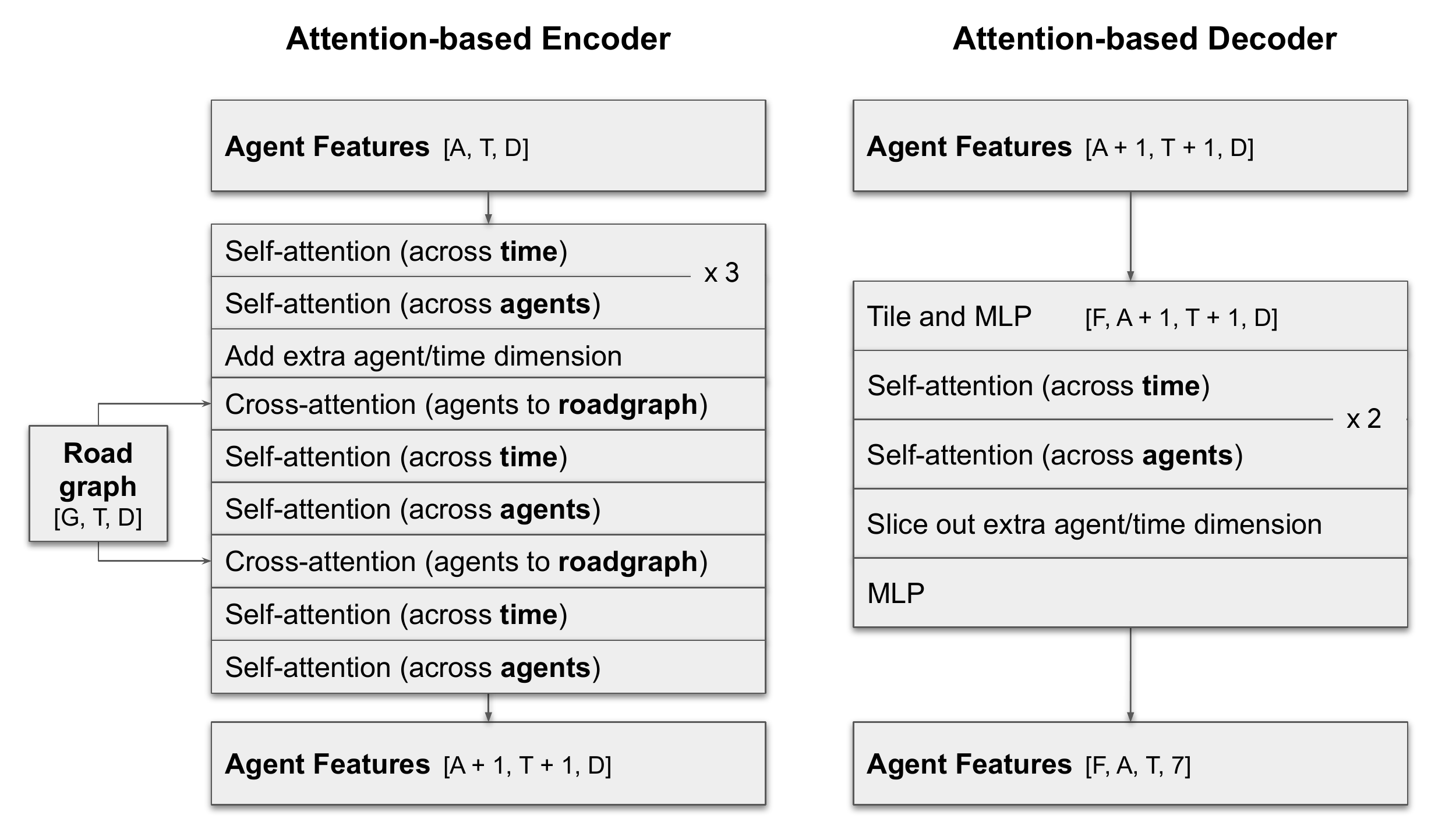} 
\caption{\textbf{Single model architecture for multiple motion prediction tasks.} Left: Different masking strategies define distinct tasks. The left column represents current time and the top row represents the agent indicating the autonomous vehicle (AV). A single model can be trained for data associated with motion prediction, conditional motion prediction, and goal-directed prediction, by matching the masking strategy to each prediction task. Right: Attention-based encoder-decoder architecture for joint scene modeling. Architecture employs factored attention along the time and agent axes to exploit the dependencies in the data, and cross-attention to inject side information.}
\label{fig:model-architecture-diagram}
%\vspace{-0.7cm}
\end{figure*}

The \modelname model has three stages: (i) Embed the agents and the road graph into a high dimensional space, (ii) Employ an attention-based network to encode the interactions between agents and the road graph, (iii) Decode multiple futures using an attention-based network. The model takes as input a feature for every agent at every time step, and also predicts an output for every agent at every time step. We employ an associated mask, where every agent time step has an associated indicator of 1 (hidden) or 0 (visible), indicating whether the input feature is hidden (i.e. removed) from the model. This approach mirrors the approach of masked-language models such as BERT \cite{devlin2018bert}. The approach is flexible, enabling us to simultaneously train a single model for {\it motion prediction (MP)} \cite{cui2019multimodal, chai2019multipath,lee2017desire,hong2019rules,casas2020spagnn,salzmann2020trajectron++,casas2020spagnn,liang2020laneGCN,wimp2020}, {\it conditional motion prediction (CMP)} \cite{wimp2020, kate_cbp, salzmann2020trajectron++} and {\it goal-conditioned prediction (GCP)} \cite{deo2020traj} simply by changing what data is shown to the model (Figure \ref{fig:model-architecture-diagram}, left). We summarize the key contributions below, and reserve details for the Appendix.

% Our design provides a high degree of flexibility in determining what the model takes as input and how it is queried, allowing it to perform multiple tasks simultaneously.
{\bf Multi-task representation.} The key representation in the model is a 3-dimensional tensor of $A$ agents with $D$ feature dimensions across $T$ time steps. At every layer within the architecture, we aim to maintain a representation of shape $[A, T, D]$, or when decoding, $[F, A, T, D]$ across $F$ potential futures. Each task (MP, CMP, GCP) can be formulated as a query with a specific masking strategy by setting the indicator mask to 0, thus providing that data to the model  (Figure \ref{fig:model-architecture-diagram}, left). The goal of the model is to impute the features for each shaded region corresponding to subsets of time and agents in the scenario that are masked. 
% Note that we additionally employ data augmentation to boost performance as detailed in the Appendix.

\subsection{Scene-centric representation for agents and road graphs.}
We use a scene-centric embedding where we use an agent of interest's position as the origin \footnote{For \wmod, we center the scene with respect to the autonomous vehicle (AV). For Argoverse, we center the scene with respect to the agent that needs to be predicted. Both are centered around what would be the last visible time step in a motion prediction setup for all tasks.}, and encode \textit{all} roadgraph and agents with respect to it. This is contrast to approaches which use an agent-centric representation, where the representations are computed separately for each agent, treating each agent in turn as the origin. 
% The scene-centric representation is crucial for enabling our model to perform joint trajectory predictions across multiple agents. 

In detail, we first generate a feature for every agent time step if that time step is visible. Second, we generate a set of features for the static road graph, road elements static in space and time, learning one feature vector {\it per polyline} (with signs being polylines of length 1) using a PointNet \cite{qi2017pointnet}. Last, we generate a set of features for the dynamic road graph, which are road elements static in space but dynamic in time (e.g. traffic lights), also one feature vector per object. All three categories have $xyz$ position information, which we preprocess to center and rotate around the agent of interest and then encode with sinusoidal position embeddings \cite{vaswani2017attention}.

\subsection{Encoding transformer}

We focus on a simple encoder-decoder attention-based architecture which maintains a representation of $[A, T, D]$ throughout (Figure \ref{fig:model-architecture-diagram}, right). We summarize the architecture briefly, but reserve details for the Appendix and Table \ref{table:architecture}. The majority of layers are a form of the Transformer layer \cite{vaswani2017attention} (Table \ref{table:transformer}). Attention layers are parameterized as matrices representing the query $Q$, key $K$, and value $V$, whose output $y={\sf softmax}\frac{\left(Q\,K^T\right)V}{\sqrt{dim_k}}$. Each matrix is computed as a learned linear transformation of the underlying representation $x$, e.g. $Q = W_q\,x$. Each attention layer is followed by a feed-forward layer of the same hidden dimension, and a skip connection addition of the result with the input to the whole Transformer layer. All layers of the encoder and decoder employ a $D$ feature dimension. The final layer after the decoder is a 2-layer MLP that predicts 7 outputs. The first 6 outputs correspond to the 3-dimensional position of an agent at a given time step in \textit{absolute coordinates} (e.g. meters) with respect to the agent of interest, and the corresponding uncertainty parameterized by a Laplace distribution \cite{meyer2020uncertainty}. The remaining dimension predicts the heading. 

{\bf Efficient factorized self-attention.}
The bulk of the computation is performed with a Transformer \cite{vaswani2017attention} (Table \ref{table:transformer}). One naive approach to use the Transformer would be to perform attention directly on the entire set of agent and time step features (i.e., attention across $AT$ dimensions). However, this approach is computationally expensive, and also suffers from an identity symmetry challenge: since we do not add any specific agent identity indicator, two agents of the same type with the same masked future time-step will have the same input representation to the transformer, resulting in the same output. Thus, we design a {\it factorized} attention based on the time and agents axes (for related ideas, see \citet{wang2020axial,szegedy2016rethinking,ho2019axial}). 

Applying attention only across time allows the model to learn smooth trajectories independent of the identity of the agent. Likewise, applying attention only across agents allows the model to learn multi-agent interactions independent of the specific time step. Finally, in order to capture both time and agent dependencies, the model simply alternates attention across agents and time in subsequent layers (Figure \ref{fig:model-architecture-diagram}, right panel). The model is also permutation equivariant to the ordering of the agents at input, since the attention operation is permutation equivariant.

{\bf Cross-attention.}
In order to exploit side information, which in our case is a road graph, we use cross-attention to enable the agent features to be updated by attending to the road graph. Concretely, we calculate the queries from the agents, but the keys and values come from the embeddings of the road graph. The road graph embeddings are final after the per-polyline PointNet, and therefore \textit{not} updated during these attention layers. This requires that the model learn interactions between the road structure and agents that are independent of the specific time step or agent. We highlight that the road graph representation is also permutation-equivariant and shared across all agents in the scene, whereas prior approaches have often used a per-agent road graph representation.

\subsection{Predicting probabilities for each futures.}

Our model also needs to predict a probability score for each future (in the joint model) or trajectory (in the marginal model). In order to do so, we need a feature representation that summarizes the scene and each agent. 
% To accomplish this, we generate an artificial agent and time step and add this to the agent features representation (Figure \ref{fig:model-architecture-diagram}, left panel).  
After the first set of factorized self-attention layers, we compute the mean of the agent features tensor across the agent and time dimension separately, and add these as an additional artificial agent and time making our internal representation  $[A+1, T+1, D]$ (Figure \ref{fig:model-architecture-diagram}, left panel). This artificial agent and time step propagates through the network, and provides the model with extra capacity for representing each agent, that is not tied to any timestep. At the final layer, we slice out the artificial agent and time step to obtain summary features for each agent (the additional time per agent), and for the scene (the `corner' feature that is both additional time and agent). This feature is then processed by a 2-layer MLP producing a single logit value that we use with a softmax classifier for a permutation equivariant estimate of probabilities for each futures.

\subsection{Joint and marginal loss formulation.}\label{methods_loss}

The output of our model is a tensor of shape $[F, A, T, 7]$ representing the location and heading of each agent at the given time step. Because the model uses a scene-centric representation for the locations through positional embeddings, the model is able to predict all agents simultaneously in a single feed-forward pass. This design also makes it possible to have a straight-forward switch between joint future predictions and marginal future predictions.

To perform joint future prediction, we treat each future (in the first dimension) to be coherent futures across all agents. Thus, we aggregate the displacement loss across all agents \footnote{Each dataset identifies a subset of agents to be predicted. We only include this subset in our loss calculation. For Argoverse, this is 1 agent; for \wmod this is 2-8 agents.} and time steps to build a loss tensor of shape $[F]$. We only back-propagate the loss through the individual future that most closely matches the ground-truth in terms of displacement loss \cite{SocialGAN,yeh2019diverse}. 
%minimum loss among all $[F]$ futures as the one that most closely match the ground-truth, and only back-propagate the error through it.
For marginal future predictions, each agent is treated independently. After computing the displacement loss of shape $[F, A]$, we do \textit{not} aggregate across agents. Instead, we select the future with minimum loss for each agent separately, and back-propagate the error correspondingly (Appendix, Figure \ref{fig:loss-code}). 
%This allows the agents to select among different futures. One notable benefit of our approach is that we are able to produce predictions for all agent simultaneously. 
% Pseudo-code is available in Appendix Figure \ref{fig:loss-code}.

{\bf Evaluation metrics for motion prediction.} We evaluate the quality of $k$ weighted trajectory hypotheses using the standard evaluation metrics: minADE, minFDE, miss rate, and mAP. Each evaluation metric attempts to measure how close the top $k$ trajectories are to ground truth observation. A simple and common distance-based metric is to measure the $L_2$ norm between a given trajectory and the ground truth \cite{alahi2016social, pellegrini2009you}.  minADE reports the $L_2$ norm of the trajectory with the minimal distance. minFDE likewise reports the $L_2$ norm of the trajectory with the smallest distance only evaluated at the final location of the trajectory. 
We additionally report the miss rate (MR) and mean average precision (mAP) to capture how well a model predicts all of the future trajectories of agents probabilistically \cite{yeh2019diverse,chang2019argoverse,ettinger2021large}.  For joint future evaluation settings, we measure the scene-level equivalents (minSADE, minSFDE, and SMR) that evaluate the prediction of the best single consistent future \cite{casas2020implicit}.
\section{Results}

We evaluate the Scene Transformer on motion prediction tasks from the Argoverse dataset \cite{chang2019argoverse} and Waymo Open Motion Dataset (WOMD) \cite{ettinger2021large}. The Argoverse dataset consists of 324,000 run segments (each 5 seconds in length) from 290 km of roadway containing 11.7 million agent tracks in total, and focuses on single agent (marginal) motion prediction. The WOMD dataset consists of 104,000 run segments (each 20 seconds in length) from 1,750 km of roadway containing 7.64 million unique agent tracks. Importantly, the WOMD has two tasks, each with their own set of evaluation metrics: a \textit{marginal motion prediction challenge} that evaluates the quality of the motion predictions independently for each agent (up to 8 per scene), and a \textit{joint motion prediction challenge} that evaluates the quality of a model's joint predictions of exactly 2 agents per scene. We train each model on a Cloud TPU \cite{jouppi2017datacenter} See Appendix for all training details. 
% We demonstrate that the resulting model achieves best published performance across these datasets.

First, in Section \ref{marginal_results}, we focus on the marginal prediction task and show that \modelname achieves competitive results on both Argoverse \cite{chang2019argoverse} and the WOMD \cite{ettinger2021large}. In Section \ref{joint_results}, we focus on the joint prediction task and train the Scene Transformer with our joint loss formulation (see Section \ref{methods_loss}). We show that with a single switch of the loss formula, we can achieve superior \textit{joint motion prediction} performance. In Section \ref{factorized-self-atten} we discuss factorized versus non-factorized attention. Finally, in Section \ref{multitask} we show how our masked sequence model formulation allows us to train a multi-task model capable of motion prediction, conditional motion prediction, and goal-conditioned prediction. In Appendix \ref{tradeoffs} we discuss the trade-off between marginal and joint models. 
% For all models, we evaluate the quality of $k$ weighted trajectory hypotheses.
% using a suite of standard evaluation metrics (See Methods). 

\begin{table}[t]
\centering
\footnotesize
%\scriptsize
\begin{tabular}{c|ccc}
    \toprule
    Method & minADE$\,\downarrow$ & minFDE$\,\downarrow$ & MR$\,\downarrow$ \\
    \midrule
     Jean \citep{mercat2020multi} & 0.97 & 1.42 & 0.13 \\
     WIMP \citep{wimp2020} & 0.90 & 1.42 & 0.17 \\
     TNT \citep{zhao2020tnt} & 0.94 & 1.54 & 0.13 \\
     LaneGCN \citep{liang2020laneGCN} & 0.87 & 1.36 & 0.16 \\
     TPCN \citep{ye2021tpcn} & 0.85 & 1.35 & 0.16 \\
     mmTransformer \citep{liu2021multimodal} & 0.84 & 1.34 & 0.15 \\
     HOME \citep{HOME2021} & 0.94 & 1.45 & {\bf 0.10} \\
     %MulitPath++ & 0.89 & 1.50 & 0.131 \\
     Ours (marginal) & {\bf 0.80} & {\bf 1.23} & 0.13 \\
     \bottomrule
\end{tabular}
\vspace{0.2cm}
\caption{\textbf{Marginal predictive performance on Argoverse motion prediction.} Results reported on {\it test} split for vehicles \cite{chang2019argoverse}. minADE, minFDE reported for $k=6$ predictions \cite{alahi2016social,pellegrini2009you}; Miss Rate (MR) \cite{chang2019argoverse} within 2 meters of the target. All results are reported for $t=3$ seconds. \label{table:argoverse-test-results}}
\vspace{-0.2cm}
\end{table}

%\input{sections/womd-table-old}
%\clearpage

% Some baseline numbers gathered from here:
% https://docs.google.com/document/d/1pKeoR0zzl21BoLjQ8fSe_nUt5hVSNueBTw1Yes5yPcA/edit
\begin{table}[t]
\centering
\footnotesize
\bgroup
\def\arraystretch{1.05}%  1 is the default.
\resizebox{\linewidth}{!}{
\begin{tabular}{cc|ccc|ccc|ccc|ccc}
    \toprule
    \multicolumn{2}{c|}{\multirow{2}{*}{{\bf Motion Prediction}}} & \multicolumn{3}{c|}{minADE$\,\downarrow$} & \multicolumn{3}{c|}{minFDE$\,\downarrow$} & \multicolumn{3}{c|}{MR$\,\downarrow$} & \multicolumn{3}{c}{mAP$\,\uparrow$} \\
    & & veh & ped & cyc & veh & ped & cyc & veh & ped & cyc & veh & ped & cyc\\ 
    \midrule
    {\it valid} & & & & & & & & & & & \\
     & LSTM baseline \cite{ettinger2021large} & 1.34 & 0.63 & 1.26 & 2.85 & 1.35 & 2.68 & 0.25 & 0.13 & 0.29 & 0.23 & 0.23 & 0.20 \\
     % From leaderboard submission
     & Ours (marginal) & {\bf 1.17} & \textbf{0.59} & \textbf{1.15} & \textbf{2.51} & \textbf{1.26} & \textbf{2.44} & \textbf{0.20} & \textbf{0.12} & \textbf{0.24} & \textbf{0.26} & \textbf{0.27} & \textbf{0.20} \\
    %  & ours (marginal + NMS) & 1.23 & 0.75 & 1.25 & 2.64 & 1.68 & 2.68 & 0.20 & 0.24 & 0.27 & \textbf{0.33} & 0.26 & \textbf{0.26} \\
     \midrule
    {\it test} & & & & & & & & & & & & \\
     & LSTM baseline \cite{ettinger2021large} & 1.34 & 0.64 & 1.29 & 2.83 & 1.35 & 2.68 & 0.24 & 0.13 & 0.29 & 0.24 & 0.22 & 0.19\\
     & ReCoAt \cite{Huang_2021} & 1.69 & 0.69 & 1.47 & 3.96 & 1.51 & 3.30 & 0.40 & 0.20 & 0.37 & 0.18 & 0.25 & 0.17 \\
     & SimpleCNNOnRaster \cite{Konev_2021} & 1.47 & 0.71 & 1.39 & 3.18 & 1.52 & 2.89 & 0.27 & 0.16 & 0.31 & 0.19 & 0.18 & 0.14 \\
     & DenseTNT \cite{Gu_2021} & 1.35 & 0.85 & 2.17 & 3.35 & 1.40 & 2.94 & 0.20 & 0.13 & 0.23 & \textbf{0.28} & \textbf{0.28} & \textbf{0.21} \\
     & Ours (marginal) & \textbf{1.17} & \textbf{0.60} & \textbf{1.17} & \textbf{2.48} & \textbf{1.25} & \textbf{2.43} & \textbf{0.19} & \textbf{0.12} & \textbf{0.22} & 0.27 & 0.23 & 0.20\\
    %  & ours (marginal + NMS) & 1.22 & 0.76 & 1.27 & 2.61 & 1.68 & 2.69 & 0.20 & 0.23 & 0.27 & \textbf{0.33} & 0.27 & \textbf{0.25}\\
     \bottomrule
\end{tabular}}
\egroup
\caption{\textbf{Marginal predictive performance on Waymo Open Motion Dataset motion prediction}. Results presented on the {\it standard} splits of the validation and test datasets \cite{ettinger2021large} evaluated with traditional \textit{marginal metrics} for $t=8$ seconds. minADE, minFDE reported for $k=6$ predictions \cite{alahi2016social,pellegrini2009you}; Miss Rate (MR) \cite{chang2019argoverse} within 2 meters of the target. See Appendix for $t=3$ or $5$ seconds. We include the challenge winner results in this table~\cite{womd_winners}. \label{table:wmod-marginal-results}}
\vspace{-0.1cm}
\end{table}

\begin{table}[t]
\centering
\footnotesize
\bgroup
\def\arraystretch{1.05}%  1 is the default.
\resizebox{\linewidth}{!}{
\begin{tabular}{cc|ccc|ccc|ccc|ccc}
    \toprule
    \multicolumn{2}{c|}{\multirow{2}{*}{{\bf Interaction Prediction}}} & \multicolumn{3}{c|}{minSADE$\,\downarrow$} & \multicolumn{3}{c|}{minSFDE$\,\downarrow$} & \multicolumn{3}{c|}{SMR$\,\downarrow$} & \multicolumn{3}{c}{mAP$\,\uparrow$}\\
    & & veh & ped & cyc & veh & ped & cyc & veh & ped & cyc & veh & ped & cyc\\ 
    \midrule
    {\it valid} & & & & & & & & & & &\\
     & LSTM baseline \cite{ettinger2021large} & 2.42 & 2.73 & 3.16 & 6.07 & 4.20 & 6.46 & 0.66 & 1.00 & 0.83 & 0.07 & \textbf{0.06} & 0.02 \\
     % From leaderboard submission
     & Ours (marginal-as-joint) & 2.04 & 1.62 & 2.28 & 4.94 & 3.81 & 5.67 & 0.54 & 0.63 & 0.72 & \textbf{0.11} & 0.05 & 0.03 \\
     % From leaderboard submission
     & Ours (joint, MP-only) & \textbf{1.72} & \textbf{1.38} & 1.96 & \textbf{3.98} & \textbf{3.11} & 4.75 & \textbf{0.49} & \textbf{0.60} & 0.73 & \textbf{0.11} & 0.05 & 0.03  \\
     % At 371.2K steps
     % https://tensorboard.corp.google.com/experiment/mldash:chauffeur:5237978701955273581/#scalars&regexInput=Dev&tagFilter=joint_metrics%2FVehicle_at_8s&_smoothingWeight=0
     & Ours (joint, multi-task) & \textbf{1.72} & 1.39 & \textbf{1.94} & 3.99 & 3.15 & \textbf{4.69} & \textbf{0.49} & 0.62 & \textbf{0.71} & \textbf{0.11} & \textbf{0.06} & \textbf{0.04} \\     
     \midrule
    {\it test} & & & & & & & & & & & \\
     & LSTM baseline \cite{ettinger2021large} & 2.46 & 2.47 & 2.96 & 6.22 & 4.30 & 6.26 & 0.67 & 0.89 & 0.89 & 0.06 & 0.03 & 0.03\\
     & HeatIRm4 \cite{Mo_2021} & 2.93 & 1.77 & 2.74 & 7.20 & 4.06 & 6.69 & 0.80 & 0.80 & 0.91 & 0.07 & \textbf{0.05} & 0.00 \\
    %  & AIR2 \cite{Wu_2021} & 2.32 & 1.69 & 2.47 & 5.00 & 3.68 & 5.47 & 0.64 & 0.71 & 0.81 & 0.10 & 0.04 & 0.02\\
     % At 857.6K steps
     & Ours (marginal-as-joint) & 2.08 & 1.62 & 2.24 & 5.04 & 3.87 & 5.41 & 0.55 & 0.64 & 0.73 & 0.08 & \textbf{0.05} & 0.03 \\
     & Ours (joint, MP-only) & 1.76 & \textbf{1.38} & \textbf{1.95} & 4.08 & \textbf{3.19} & \textbf{4.65} & \textbf{0.50} & \textbf{0.62} & \textbf{0.70} & 0.10 & \textbf{0.05} & \textbf{0.04}\\
     & Ours (joint, multi-task) & \textbf{1.74} & 1.41 & \textbf{1.95} & \textbf{4.06} & 3.26 & 4.68 & \textbf{0.50} & 0.64 & 0.71 & \textbf{0.13} & 0.04 & 0.03\\
     \bottomrule
\end{tabular}}
\egroup
%\vspace{0.1cm}
\caption{\textbf{Joint predictive performance on Waymo Open Motion Dataset motion prediction}. Results presented on the {\it interactive} splits of the validation and test datasets \cite{ettinger2021large} evaluated with scene-level \textit{joint metrics} for $t=8$ seconds. minSADE, minSFDE for $k=6$ predictions \cite{alahi2016social,pellegrini2009you}; Miss Rate (MR) \cite{chang2019argoverse} within 2 meters of the target. ``\textbf{S}'' indicates a scene-level joint metric. See Appendix for $t=3$ or $5$ seconds. We include the challenge winner results in this table~\cite{womd_winners}. \label{table:wmod-joint-results}}
\vspace{-0.3cm}
\end{table}

\subsection{Marginal motion prediction}\label{marginal_results}

We first evaluate the performance of Scene Transformer trained and evaluated as a traditional marginal, per-agent motion prediction model. This is analogous to the problem illustrated in Figure \ref{figure:motivation} (left). For all results until Section \ref{multitask}, we use a masking strategy that provides the model with all agents as input, but with their futures hidden. We also mask out future traffic light information. 

{\bf Argoverse.} 
% We start by validating a simplified version of \modelname for the behavior prediction task on the Argoverse dataset. We specify the task of behavior prediction by masking out the future positions of agents in the scene. 
We evaluate on the popular Argoverse \cite{chang2019argoverse} benchmark to demonstrate the efficacy of our architecture. During training and evaluation, the model is only required to predict the future of the single agent of interest. Our best Argoverse model uses $D=512$ feature dimensions and label smoothing for trajectory classification.  Our model achieves state-of-the-art results compared to published, prior work \footnote{We exclude comparing to public leaderboard entries that have not been published since their details are not available, but note that our results are competitive on the leaderboard as of the submission date.} in terms of minADE and minFDE (Table \ref{table:argoverse-test-results}).

{\bf Waymo Open Motion Dataset (\wmod).} We next evaluate the performance of our model with $D=256$ on the recently released WOMD \cite{ettinger2021large} for the marginal motion prediction task. This task is a standard motion prediction task where up to 8 agents per scene are selected to have their top 6 motion predictions evaluated independently. Our model trained with the marginal loss achieves state-of-the-art results on the minADE, minFDE, and miss rate metrics (Table \ref{table:wmod-marginal-results}). 

% In Table \ref{table:wmod-marginal-results} In the top half of Table \ref{table:wmod-results} (see "ours (marginal)") we show the validation and test set performance of variants of our model on th dataset's marginal metrics. 
% We also evaluate the same model with non-maximum suppression added (see "ours (marginal + NMS)") \footnote{We use NMS to post-process the $k=6$ output trajectories to reduce redundancy. This was important for the AP metric on WOMD, and further details are available in the supplementary material. } to remove duplicate predictions, as the new mean Average Precision (mAP) metric averages over 8 trajectory shape categories, one of which is stationary, and penalizes for duplicates. We find there exists a trade-off when using NMS between minADE/minFDE style metrics and the new mAP metric. Both variants of our marginal model achieve state-of-the-art performance over the provided baseline (Table \ref{table:wmod-marginal-results}) on the {\it Motion Prediction} challenge, further confirming the strength of our general architecture.

\subsection{Joint motion prediction}\label{joint_results}

% As shown in Section \ref{methods_loss}, we can employ a simple loss formula change to convert our model from a marginal to a joint model. This is made possible by the fact that our model already has all agents and roadgraph features in a shared global coordinate system. 
To evaluate the effectiveness of Scene Transformer when trained with a joint loss formulation (Section \ref{methods_loss}), we evaluate our model on the \textit{Interaction Prediction} challenge in WOMD \cite{ettinger2021large}. This task measures the performance of the model at predicting the \textit{joint} future trajectories of two interacting agents (Figure \ref{figure:motivation}, right), and employs \textit{joint variants} of the common minADE, minFDE, and Miss Rate (MR) metrics denoted as minSADE, minSFDE, SMR. Note that the ``\textbf{S}'' indicates ``scene-level'' metrics. These metrics aim to measure the quality and consistency of the two agents joint prediction - for example, the joint variant of Miss Rate (SMR) only records a "hit" if both interacting agent's predicted trajectories are within the threshold of their respective ground truths. 

We find that for the \textit{Interaction Prediction} challenge our joint model's joint predictions easily outperforms the \wmod provided baseline as well as a marginal version of the model converted into a joint prediction \footnote{Note that the output of a joint model can be directly used in a marginal evaluation. However, converting the output of a marginal model into a joint evaluation is nuanced because there lacks an association of futures across agents (Figure \ref{figure:motivation}). We employ a simple heuristic to convert the outputs of a marginal model for joint evaluation: we take the top 6 pairs of trajectories from the combination of both agent's trajectories for 36 total pairs, and retain the top 6 pairs with the highest product of probabilities.} into joint predictions. (Table \ref{table:wmod-joint-results}).
%\footnote{The output of a joint model can be directly used in a marginal evaluation. However, using the output of a marginal model in a joint evaluation is extremely nuanced because there lacks an association of futures across agents (Figure \ref{figure:motivation}). We employ a simple heuristic to convert the outputs of a marginal model for joint evaluation: we take the top 6 pairs of trajectories from the combination of both agent's trajectories for 36 total pairs, and then keep the top 6 pairs with the highest combined probabilities.} into joint predictions (see "ours (marginal)").
This shows that beyond the strength of our overall architecture and approach, that explicitly training a model as a joint model significantly improves joint performance on joint metrics. A notable observation is that even though the \textit{Interaction Prediction} task only requires predicting the joint trajectories of two agents, our method is fully general and predicts joint consistent futures of all agents.

% Dataset averages over motion forecasting metrics like min(S)ADE do not capture the types of fine-grained edge cases that are important for measuring forward progress in the field.  Unfortunately, the \wmod Interactive Prediction set does not contain pre-defined metadata necessary to evaluate specific interesting slices using different models.  In Appendix X we provide some initial analysis of slices that we can compute from the \wmod, which illustrate how joint and marginal model predictions behavior differently as the scene a) scales with more agents in complicated scenes, b) differs in the average speed of agents.  We also provide results on how joint prediction models demonstrate lower *inter-prediction overlap* compared to marginal models, which we believe to be an important metric to measure going forward.

% \todo{bencaine: We don't talk about joint-as-marginal here or above, or the multi-task model. update: this is now a long footnote.}

\subsection{Factorized agents self-attention}\label{factorized-self-atten}

Factorized self-attention confers two benefits to our model: (a) it is more efficient since the attention is over a smaller set, and (b) it provides an implicit identity to each agent during the attention across time. We ran an experiment where we replaced each axis-factorized attention layer (each pair of time and agent factorized layer) with a non-axis factored attention layer. This increased the computational cost of the model and performed worse on the Argoverse validation dataset: the factorized version achieved a minADE of 0.609 with the factored version, and 0.639 with the non-factorized version. 

% TODO(anyone): Rename this something to do with counterfactual reasoning?
\subsection{Advantages of a masked sequence modeling strategy}\label{multitask}

\begin{figure*}[t]
\centering
\begin{tabular}{p{0.3\textwidth}p{0.3\textwidth}p{0.3\textwidth}}
%\textit{Scenario A} & \textit{Scenario B} & \textit{Scenario C} & \textit{Scenario D} \\
\textit{Scenario A} & \textit{Scenario B} & \textit{Scenario C}\\
\end{tabular}

\includegraphics[width=\textwidth]{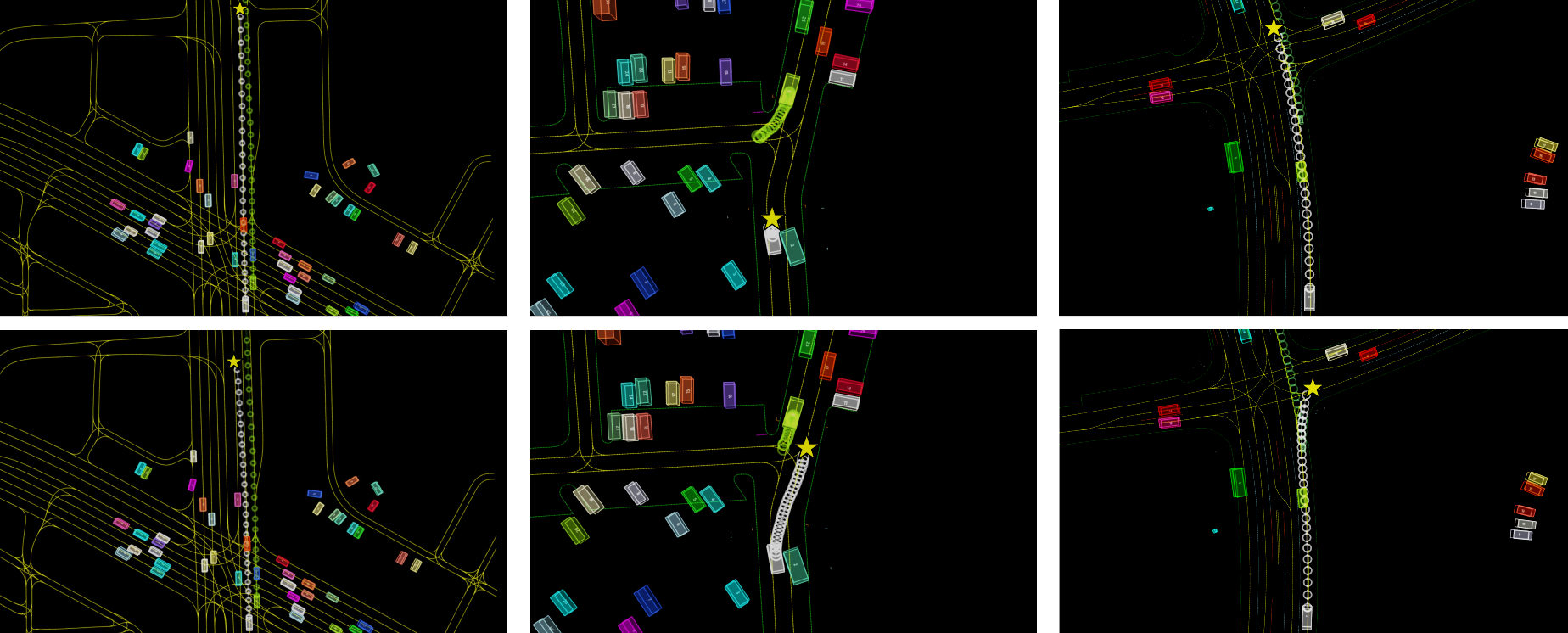}

\caption{\textbf{Goal-conditioned prediction navigates AV to selected goal positions.} Rectangles indicate vehicles on the road. Lines indicate the road graph (RG) colored by the type of lane marker. Circles indicate predicted trajectory for each agent. Star indicates selected goal for AV. Each column represents a scenario in which the AV is directed to perform one of two actions taking into account the dynamics of other agents. (A) AV instructed to either change lanes or remain in the same lane. (B) AV instructed to stop to allow oncoming vehicle to turn left, or adjust to the side in a narrow passage so that the oncoming vehicle has enough space to pass. (C) AV instructed to either proceed straight or turn right at the intersection. \label{fig:planning-visualizations}}
\end{figure*}

Our model is formulated as a masked sequence model \cite{devlin2018bert}, where at training and inference time we specify which agent timesteps to mask from the model.
This formulation allows us to select which information about any agent at any timestep to supply to the model, and measure how the model exploits or responds to this additional information.
%would condition the model's predictions.
%This allows us to provide information about any agent at any timestep to the model, and see how this additional information would condition the model's predictions.
We can express several motion prediction tasks at inference time in mask space (Figure \ref{fig:model-architecture-diagram}), in effect providing a multi-task model. We can use this unique capability to query the models for counterfactuals. What would the model predict given a subset of agents full trajectories (conditional motion prediction), or given a subset of agents final goals (goal-conditioned motion prediction)? This feature could be particularly useful for autonomous vehicle planning 
%when using the joint model, allowing us 
to predict what various future rollouts of the scene would look like given a desired goal for the autonomous vehicle (Figure \ref{fig:planning-visualizations}).

In the previous results, we used a prefix-mask during training and inference, which shows the first few steps 
% 11 timesteps for Waymo, and 21 timesteps for Argoverse) 
and predicts the remaining timesteps
% (80 timesteps for Waymo, 30 for Argoverse) 
for all agents. For this experiment, we employ two masking strategies to test on the interactive split of the \wmod, namely "conditional motion prediction" (CMP) where we show one of the two interacting agents' full trajectory, and "goal-conditioned motion prediction" (GCP) where we show the autonomous vehicle's desired goal state (Figure \ref{fig:model-architecture-diagram}). We train a model using each strategy (including MP masking) $1/3$ of the time, and evaluate the model on each all tasks. We find that across the 44k evaluation segments, the multi-task model matches the performance of our MP-only trained joint model on both joint (Table \ref{table:wmod-joint-results}) and marginal metrics (see also Appendix, Table \ref{table:wmod-results-marginal-t8}); multi-tasking training does not significantly degrade standard motion prediction performance.
% For the conditional motion prediction task, we find showing one of the two interacting agent's full trajectory \footnote{We run evaluation where we show Agent A and predict agent B, and then repeat this with the agents flipped and average the overall results.} reduces the overall minADE (for non-autonomous vehicle agents) on the validation set by 0.09 (from 1.34 to 1.25), showing that on average the model can take advantage of this extra information. Finally,  we measure the change in autonomous vehicle minADE when we condition on the goal state, showing a drop in minADE from 0.95 to 0.60, again showing the model can use this additional information. Additional analysis of these masking strategies can be found in the Appendix.
%With the additional multi-task training, the resulting model may now condition on side information. As a first test, we measure  in the CMP and GCP setting that the model reduces the overall minADE of the prediction (For CMP, 1.34 $\rightarrow$ 1.25 for non-AV; For GCP, 0.95 $\rightarrow$ 0.60 for AV). As a second test, 
We qualitatively examine the performance of the multi-task model in a GCP setting, and observe that the joint motion predictions for the AV and non-AV agents flexibly adapt to selected goal points for the AV (Figure \ref{fig:planning-visualizations}). 
%Although no quantitative benchmarks are available in the community for quantifying GCP predictions, we take these results as positive indication that the model is responding appropriately and save further exploration of counterfactual analysis for future work.

We note that dataset average metrics like min(S)ADE do not capture forecasting subtleties~\cite{ivanovic2021metrics} that are critical for measuring progress in the field.  Unfortunately, the Waymo Motion Open Dataset Interactive Prediction set does not contain pre-defined metadata necessary to evaluate specific interesting slices using different models. In Appendix~\ref{section:slicing-results} we provide some initial analysis of slices that we could compute on the dataset, which illustrate how joint and marginal model predictions behavior differently as the scene a) scales with more agents in complicated scenes, b) differs in the average speed of agents. We also show that joint prediction models demonstrate \emph{joint consistent futures} via lower inter-prediction overlap compared to marginal models.
\section{Discussion}

%Navigating dynamic environments necessitates predicting the interactions of multiple agents and the cascading effects of each potential future. Such a problem is particularly acute in the case of self-driving cars where agents and their associated behaviors may be diverse, and the decisions of the self-driving itself may significantly influence the environment. Artificially differentiating the problem into predicting the behavior of every other agent and that of the autonomous vehicle generates additional uncertainty about an agent’s behavior, because structuring the problem as such converts what may be a narrow conditional probability distribution in to a broad marginal probability distribution which accounts for all possible outcomes given interacting agents. 

We propose a unified architecture for autonomous driving that is able to model the complex interactions of agents in the environment. Our approach enables a single model to perform motion prediction, conditional motion prediction, and goal-conditioned prediction. 
% This architecture may be trained on the union of datasets tailored to individual tasks, and benefit from learning shared representations across all of these tasks. 
% Importantly, our formulation is heavily inspired by recent developments in language modeling that demonstrate that pretraining a large model on a corpus provides a powerful representation that may be exploited for few-shot learning by constructing a query on the fly \cite{brown2020language,devlin2018bert}. Similarly, in our work we construct a query by correspondingly constructing a masking strategy that matches the given autonomous vehicle task. We find that the resulting model may perform competitively if not state-of-the-art performance on an array of behavior prediction tasks. 
In the field of autonomous driving, elaborations of this problem formulation may result in learning models for planning systems that quantitatively improve upon existing systems \cite{buehler2009darpa,montemerlo2008junior,ziegler2014trajectory,zeng2019end,liu2021deep}. Likewise, such modeling efforts may be used to directly go after issues of identifying interacting agents in an environment, and potentially provide an important task for identifying causal relationships \cite{arjovsky2019invariant,scholkopf2021toward}. 
% More broadly, this work may provide an impetus to continue efforts in developing large pre-trained models paired with few-shot learning to solve multiple, related tasks. Immediate applications may arise in fields such as multimodal learning where large corpora exist for each modality and large opportunities remain for efficiently learning how these datasets may be effectively leveraged to result in synergistic benefits \cite{vinyals2015show,ramesh2021zero}.

% Restore for final submission.
%\input{sections/acknowledgements}
%\input{sections/contributions}

\bibliography{paper}
\bibliographystyle{iclr2022_conference}

\clearpage
\appendix

{\bf \Large Appendix}

\section{Architecture and Training Details}

{\bf Overview.} Table \ref{table:architecture} provides additional details about the \modelname architecture and training. The bulk of the computation and parameters reside in the Transformer layers. Table \ref{table:transformer} lists all of the operations and learned parameters in our implementation of a Transformer. MLPs are employed to embed the original data (e.g, layers $A$, $B$, $C$), build $F$ futures from the encoding (layer $T$), and predict the final position, headings and uncertainties of all agents (layers $Z_1$ and $Z_2$). The resulting attention-based architecture using $D=256$ feature dimensions contains 15,296,136 parameters.  Nearly all settings used for both \wmod and Argoverse are identical, except that for Argoverse we use $D=512$ and label smoothing to get our best marginal results.  The resulting model is trained on a TPU custom hardware accelerator \cite{jouppi2017datacenter} and converges in about 3 days of training.

{\bf Decoding multiple futures.} In order to allow the decoder to output $F$ distinct futures, we perform the following operations.
The decoder receives as input a tensor of shape $[A, T, D]$ corresponding to $A$ agents across $T$ time steps and $D$ feature dimensions. The following series of simple operations restructures the representation to predict $F$ futures. First, the representation is tiled $F$ times to generate $[F, A, T, D]$. We append a one-hot encoding to the final dimension where a $1$ indicates which the identity of each future, resulting in a tensor of shape $[F, A, T, D + F]$. The one-hot encoding allows the network to learn an embedding for each of the $F$ futures. For computational simplicity, the resulting representation is propagated through a small MLP to produce a return a tensor of the shape $[F, A, T, D]$.

{\bf Padding and Hidden Masks.} Padding and hidden masks are important to get right in the implementation of such a model. In particular, we need to ensure that the masks do not convey additional future information to the model (e.g., if the model knows which timesteps an agent is visible or occluded based on the padding in the data, it may take advantage of this and not be able to generalize). We use the concept of padding to indicate the positions where the \textit{input} is absent or provided. This is distinct from the hidden mask of shape $[A, T]$ that is used for \textit{task specification}. The hidden mask is used to query the model to inform it on which locations to predict, while the padding tells us which locations have inputs and ground-truth to compute a loss over. All padded positions are set to be hidden during preprocessing, regardless of the masking scheme, so the model tries to predict their values.
% Concretely, we use a padding value of 1 to indicate the data was hidden (and features zero'd out), and a value of 0 indicating inputs are provided for that position. 
% Our model handles padding in the same manner popularly employed by language models, having a 0/1 indicator for each agent time step (0 means the time step is valid, 1 that it is padded) also of shape $[A, T]$. 
All layers in our implementation are padding aware, including normalization layers (like batch normalization used in our encoding MLPs) and attention operations. Our attention operations set the attention weights to 0 for padded elements. If after providing our task specific hidden mask, no non-padded (valid) time steps exist for an agent, the whole agent is set to be padded. This prevents agent slots with no valid data from being used in the model. 

{\bf Predicting uncertainties.} For each predicted variate such as position or heading, we additionally predict a corresponding uncertainty. In early experiments we found that predicting a paired uncertainty improves the predictive performance of the original variate and provides a useful signal for interpreting the fidelity of the predicted value. We employ a loss function that predicts a parameterization of the uncertainty corresponding to a Laplace distribution \cite{meyer2020uncertainty}. 

{\bf Loss in lateral and longitudinal coordinates.} We use a Laplace distribution parameterization for the loss, with a diagonal covariance such that each axis is independent. To enable the model to learn meaningful uncertainties, we rotate the scene per box prediction such that each prediction's associated ground-truth is axis aligned. This formulation results in uncertainty estimates that correspond to the agent's lateral and longitudinal coordinates.

{\bf Heading representation:} Our output format is 7 dimensional, where one of the dimensions, heading, is represented in radians.  To supervise the loss on the heading dimension, we employ a now standard method of "wrapping" the angle difference between the predicted and groundtruth heading for each agent; see Figure~\ref{fig:angle}. 

\begin{figure}
\centering
\begin{minted}[fontsize=\footnotesize]{python}
def WrapAngleRadians(angles_rad, min_val=-np.pi, max_val=np.pi):
  max_min_diff = max_val - min_val
  return min_val + tf.math.floormod(angles_rad + max_val, max_min_diff)
  
heading_diff = pred_heading - gt_heading
heading_error = WrapAngleRadians(heading_diff, min_val=-PI, max_val=PI)
heading_loss = HuberLoss(heading_error)
\end{minted}
\caption{Pseudo-code for heading loss component.}
\label{fig:angle}
\end{figure}

%Finally this representation is supplied to 4 alternating factorized attention layers across agents and time. The final output of the decoder in turn results from a small MLP applied to the output with 6 target features dimensions corresponding to the 3-dimensional position and the corresponding uncertainty as parameterized with a Dirichlet distribution \cite{malinin2019reverse}. \todo{Note that the MLP contains no batch normalization and employs a ReLU nonlinearity.}

{\bf Marginal predictions from a joint model.} The model has the added feature of being easily adapted for making marginal predictions for each agent. To produce per-trajectory scores for each agents predictions, we attach an artificial extra time step to the end of the $[A, T, D]$ agent feature matrix to give us a matrix of shape $[A, T+1, D]$. This additional feature provides the model with extra capacity for representing each agent, that is not tied to any timestep. This additional feature can be used to predict which of the $k$ trajectories most closely matches the ground truth for that agent. 
% Note (from bencaine@): Not sure we want to use the word calibrated, since these scores are definitely not calibrated in any sense.
%We find it useful to likewise attach a calibrated score for each agent prediction. In order to ensure this feature, we attach an artificial time step to the end of the trajectory. For this final time step, we employ this featurization to provide a per-agent representation that may predict which of the $k$ trajectories most closely matches ground truth. The resulting classifier provides a calibrated score that may be employed for any analysis that depends on calibrated score.

{\bf Data augmentation.} We use two methods for data augmentation to aid generalization and combat overfitting on both datasets. First, we use agent dropout, where we artificially remove non-predicted agents with some probability. We found a probability of $0.1$ worked well for both datasets. We also found it beneficial to randomly rotate the entire scene between $[-\frac{\pi}{2}, \frac{\pi}{2}]$ after centering to the agent of interest. On Argoverse, this agent of interest is the agent the task designates to be predicted, where as on the \wmod we always center around the autonomous vehicle (AV), even though for some tasks it is not one of the predicted agents. Lastly, each Argoverse scene contains many agents that the model is not required to predict; we employ these contextual agents as additional target training data if the contextual agents moved by at least 6m.   On \wmod, Table~\ref{table:aug} shows data augmentation on the standard validation set does improve minADE modestly, though does not account for the majority of the benefit of the model.

\begin{table}
\centering
\footnotesize
%\scriptsize
\begin{tabular}{c|ccc}
    \toprule
    With Augmentation? & Vehicle minADE @8s & Ped minADE @8s & Cyclist minADE @ 8s \\
    \midrule
     No & 1.30 & 0.69 & 1.29 \\
     Yes & 1.18 & 0.59 & 1.15 \\
     \bottomrule
\end{tabular}
\caption{\wmod validation set performance with and without data augmentation during training. \label{table:aug}}
%\vspace{0.2cm}
\end{table}

%With Augmentation?	Vehicle minADE @ 8s	Pedestrian minADE @ 8s	Cyclist minADE @ 8s
%No	1.298	0.687	1.291
%Yes	1.178	0.592	1.153

{\bf Argoverse classification label smoothing:} Our best model on Argoverse uses $D=512$ feature dimensions, but naively scaling the model this way leads to severe overfitting on the classification subtask.  During training of our best model we employ label smoothing (0.1 + 1/6 for negatives and 0.9 + 1/6 for positive target).

{\bf WOMD redundant trajectory combination:} The AP evaluation metric on the WOMD expects that exactly one of the trajectories is given a high probability and the other trajectories a low probability. For example, the future prediction for a stationary agent is expected to have only one future with a high probability score, with a low score for the rest -- even though the trajectories may all be the same, in this case, stationary. Our best model combines redundant trajectories together if they are close in spatial location (less than $3.2$m) at the final timestep. 

% {\bf Measuring agent inter-prediction overlap.} We measure \emph{predicted} agent overlap on the best set of trajectory produced by the model. For the joint model, this corresponds to the joint prediction that has the higher probability score. For the marginal model, we take the top scoring trajectory for each agent to obtain the best set. For every predicted agent in the trajectory, we determine if it has an overlap with any \textit{other predicted} agent by comparing the rotated upright 3D bounding boxes using our predicted xyz and heading. The inter-prediction overlap rate is the number of predicted agents that are involved in some overlap, divided by the total number of predicted agents. We count two agents as overlapping if the intersection over the union (IoU) exceeds 0.01.

{\bf Embedding of agents and road graph.} To generate input features, we use sinusoidal positional embeddings \cite{vaswani2017attention} to embed the time (for agents and dynamic roadgraph) and $xyz$-coordinates separately into a $D$ dimensional features per dimension. We encode the type of each object using a one-hot encoding (e.g. object type, lane type, etc), and concatenate any other features provided in the data set such as yaw, width, length, height, and velocity. Dynamic road graphs have a second one-hot encoding indicating state, like the traffic light state. Lastly, for agents and the dynamic road graph, we add a binary indicator on whether the agent is hidden at that time step. If the agent or dynamic road graph element is hidden, all input features (e.g. position, type, velocity, state, etc) are set to 0 before encoding except the time embedding, which are linearly spaced values at the dataset's update rate starting at 0, and the hidden indicator.

For agents and the dynamic road graph, we use a 2 layer MLP with a hidden and output dimension of $D$ to produce a final feature per agent or object and per time step. For the static road graph, we must reduce a point cloud of up to $20{,}000$ road graph points, each belonging to a smaller set of polylines, to a single vector per polyline. Because some lanes can be extremely long, we break up any polyline longer than 20 points into a new set of smaller polylines. Then, we apply a small PointNet \cite{qi2017pointnet} architecture with a 2 layer MLP with a hidden and output dimension of $D$ to each point, and use max pooling per polyline to get a final feature per element.

{\bf Inference latency.} \modelname produces a prediction for every agent in the scene in a single pass. Its inference speed depends on the number of agents in the scene. Our preliminary profiling of inference speed ranged from 52 ms (32 agents) to 175 ms (128 agents) on the Waymo Open Motion Dataset (128 agents, 91 timesteps for each agent, up to 1400 roadgraph elements). This was measured on an Nvidia V100 using a standard TensorFlow inference graph in float32 without optimization tricks, optimization tuning, etc.  This is in line with the expected linear scaling of the factorized attention modules. 

% /cns/oi-d/home/chauffeur/ml/tensorflow/bencaine/babelfish/rs=6.3/ttl=90d/waymo_mod_head_04_29_21.v0/train_train/model_analysis.txt
\begin{table*}[t]
%\scriptsize
%\small
\tiny
\centering
\bgroup
\def\arraystretch{1.2}%  1 is the default, change whatever you need
\begin{tabular}{lccccccccr}
\toprule
Meta-Arch & Name & Input & Operation & Queries & Keys/Values & Across & Atten Matrix & Output Size & \# Param \\
\midrule
{\bf Encoder} \\
%{\bf Embedder} \\
& $\mathcal{A}$ & Agents &  MLP + BN & -- & -- & -- & -- & $[A, T, D]$ & 334080 \\
& $\mathcal{B}$ & Dyna RG & MLP + BN & -- & -- & -- & -- & $[G_D, T, D]$ & 337408 \\
& $\mathcal{C}$ & Static RG &MLP + BN & -- & -- & -- & -- & $[G_S, T, D]$ &  270592 \\
\\
& $\mathcal{D}$ & $\mathcal{A,B,C}$ &  Transformer & Agents & Agents & Time & $[A, T, T]$ & $[A, T, D]$ & 789824 \\
& $\mathcal{E}$ & $\mathcal{D}$ & Transformer & Agents & Agents & Agents & $[T, A, A]$ & $[A, T, D]$ & 789824 \\
\\
& $\mathcal{F}$ & $\mathcal{E}$ & Transformer & Agents & Agents & Time & $[A, T, T]$ & $[A, T, D]$ & 789824 \\
& $\mathcal{G}$ & $\mathcal{F}$ & Transformer & Agents & Agents & Agents & $[T, A, A]$  & $[A, T, D]$ &  789824 \\
\\
& $\mathcal{H}$ & $\mathcal{G}$ & Transformer & Agents & Agents & Time & $[A, T, T]$ & $[A, T, D]$ & 789824 \\
& $\mathcal{I}$ & $\mathcal{H}$ & Transformer & Agents & Agents & Agents & $[T, A, A]$ & $[A, T, D]$ &  789824 \\
\\
& $\mathcal{J}$ & $\mathcal{I}$ & Transformer & Agents & Static RG & Time & $[T, A, G_S]$ & $[A, T, D]$ &  789824 \\
& $\mathcal{K}$ & $\mathcal{J}$ & Transformer & Agents & Dyna RG & Time & $[T, A, G_D]$ & $[A, T, D]$ &  789824 \\
\\
& $\mathcal{L}$ & $\mathcal{K}$ & Transformer & Agents & Agents & Time & $[A, T, T]$ & $[A, T, D]$ & 789824 \\
& $\mathcal{M}$ & $\mathcal{L}$ & Transformer & Agents & Agents & Agents & $[T, A, A]$ & $[A, T, D]$ &  789824 \\
\\
& $\mathcal{N}$ & $\mathcal{M}$ & Transformer & Agents & Static RG & Time & $[T, A, G_S]$ & $[A, T, D]$ &  789824 \\
& $\mathcal{O}$ & $\mathcal{N}$ & Transformer & Agents & Dyna RG & Time & $[T, A, G_D]$ & $[A, T, D]$ &  789824 \\
\\
& $\mathcal{P}$ & $\mathcal{O}$ & Transformer & Agents & Agents & Time & $[A, T, T]$ & $[A, T, D]$ & 789824 \\
& $\mathcal{Q}$ & $\mathcal{P}$ & Transformer & Agents & Agents & Agents & $[T, A, A]$ & $[A, T, D]$ &  789824 \\
\\

\midrule
{\bf Decoder} \\
& $\mathcal{R}$ & $\mathcal{Q}$ & Tile & -- & -- & -- & -- & $[F, A, T, D]$ &  0 \\
& $\mathcal{S}$ & $\mathcal{R}$ & Concat & -- & -- & -- & -- & $[F, A, T, D${\tiny +}$F]$ &  0 \\
& $\mathcal{T}$ & $\mathcal{S}$ & MLP & -- & -- & -- & -- & $[F, A, T, D]$ &  68096 \\
\\
& $\mathcal{U}$ & $\mathcal{T}$ & Transformer & Agents & Agents & Time & $[A, T, T]$ & $[F, A, T, D]$ & 789824 \\
& $\mathcal{V}$ & $\mathcal{U}$ & Transformer & Agents & Agents & Agents & $[T, A, A]$  &  $[F, A, T, D]$ &  789824 \\
\\
& $\mathcal{W}$ & $\mathcal{V}$ & Transformer & Agents & Agents & Time &  $[A, T, T]$ & $[F, A, T, D]$ &  789824 \\
& $\mathcal{X}$ & $\mathcal{W}$ & Transformer & Agents & Agents & Agents & $[T, A, A]$  & $[F, A, T, D]$ &   789824 \\
\\

& $\mathcal{Y}$ & $\mathcal{X}$ & Layer Norm & -- & -- & -- & -- & $[F, A, T, D]$ & 512 \\
\\
& $\mathcal{Z}_1$ & $\mathcal{Y}$ & MLP + BN & -- & -- & -- &  -- & $[F, A, T, 6]$ & 66817 \\
& $\mathcal{Z}_2$ & $\mathcal{Y}$ & MLP & -- & -- & -- &  -- & $[F, A, T, 7]$ & 1799 \\
\\
%{\bf Total} & & & & & & & 15,296,136 \\
\midrule
% https://source.corp.google.com/piper///depot/google3/waymo/shared/brain/behavior_understanding/params/waymo_mod.py;l=254
% /cns/oi-d/home/chauffeur/ml/tensorflow/bencaine/babelfish/rs=6.3/ttl=90d/waymo_mod_head_04_29_21.v0/train/trainer_params.txt
\multicolumn{3}{l}{Optimizer} & \multicolumn{4}{l}{Adam ($\alpha = 1\mathrm{e}{-4}$, $\beta_1 = 0.9$, $\beta_2 = 0.999$)}  \\
\multicolumn{3}{l}{Learning Rate Schedule} & \multicolumn{6}{l}{Total epochs: 150; Linear ramp-up: 0.1 epochs} \\
\multicolumn{3}{l}{Batch size} & \multicolumn{6}{l}{64} \\
\multicolumn{3}{l}{Gradient Clipping (norm)} & \multicolumn{6}{l}{5.0} \\
\multicolumn{3}{l}{Weight initialization}  & \multicolumn{6}{l}{\citet{glorot2010understanding}}  \\
\multicolumn{3}{l}{Weight decay}  & \multicolumn{6}{l}{None} \\
\multicolumn{3}{l}{Position Embeddings}  & \multicolumn{6}{l}{Min Timescale: 4; Max Timescale: 256} \\
\multicolumn{3}{l}{Temporal Embeddings}  & \multicolumn{6}{l}{Min Timescale: 6; Max Timescale: 80} \\
\multicolumn{3}{l}{Future classification weight}  & \multicolumn{6}{l}{0.1} \\
\multicolumn{3}{l}{Position classification weight}  & \multicolumn{6}{l}{1.0} \\
\multicolumn{3}{l}{Laplace Target Scale}  & \multicolumn{6}{l}{1.0} \\
\bottomrule

\end{tabular}
\egroup
%\vspace{0.2cm}
\caption{{\bf \modelname\,architecture and training details.} The network receives as input of $A$ agents across $T$ time steps and $K$ features. $K$ is the total number of input features (e.g. 3-D position, velocity, object type, bounding box size).  A subset of these inputs are masked. $G_S$ and $G_D$ is the maximum number of road graph (RG) elements and $D$ is the total number of features. MLP and BN denote multilayer perception and batch normalization \cite{ioffe2015batch}, respectively. The output of the network is $\mathcal{Z}_1$ and $\mathcal{Z}_2$. $\mathcal{Z}_1$ corresponds to predicting the logits for classifying which one of the $F$ futures is most likely. $\mathcal{Z}_2$ corresponds to the predicted $xyz$-coordinates with their associated uncertainties, and a single value for heading. In our model $D=256$, $K=7$ and $F=6$ for a total of 15,296,136 parameters for both datasets. For the Waymo Open Motion Dataset $G_s=1400$, $G_d=16$, $T=91$, $A=128$, and for Argoverse $G_s=256$, $G_d=0$, $T=50$, and $A=64$. All layers employ ReLU nonlinearities.}
% The top and bottom sections corresponds to the encoder and decoder, respectively.
\label{table:architecture}
\end{table*}

\begin{table*}[t]
%\scriptsize
%\footnotesize
\centering
\bgroup
\def\arraystretch{1.2}%  1 is the default, change whatever you need
\begin{tabular}[width\=linewidth]{cccccr}
\toprule
Name & Input & Operation & Parameter Sizes & Output Size & \# Param \\
\midrule
$X$ & $X_o$ & Layer Norm & $[D]$, $[D]$ & $[A, T, D]$ &  512 \\
$K$ & $X$ & Affine Projection & $[D, H, \frac{D}{H}]$, $[H, \frac{D}{H}]$ & $[A, T, H, \frac{D}{H}]$ &  65792 \\
$V$ & $X$ & Affine Projection & $[D, H, \frac{D}{H}]$, $[H, \frac{D}{H}]$ & $[A, T, H, \frac{D}{H}]$ &  65792 \\
$Q_o$ & $X$ & Affine Projection & $[D, H, \frac{D}{H}]$, $[H, \frac{D}{H}]$ & $[A, T, H, \frac{D}{H}]$ &  65792 \\
$Q$ & $Q_o$ & Rescale & $[\frac{D}{H}]$ & $[A, T, H, \frac{D}{H}]$ & 64 \\
$Y_1$ & $Q,K,V$ & $\mathsf{softmax}(Q\,K^T)V$ & -- & $[A, T, H, \frac{D}{H}]$ &  0 \\
$Y_2$ & $Y_1$ & Affine Projection & $[H, \frac{D}{H}, D]$, $[D]$ & $[A, T, D]$ &  65792 \\
$S$ & $F_1,X_o$ & Sum & --  & $[A, T, D]$ &  0 \\
$F_1$ & $Y_2$ & MLP & $[D, kD]$, $[kD]$ & $[A, T, kD]$ &  263168 \\
$F_2$ & $S$ & MLP  & $[kD, D]$, $[D]$ & $[A, T, D]$ &  262400 \\
$Z$ & $F_2$ & Layer Norm & $[D]$, $[D]$ & $[A, T, D]$ &  512 \\
\midrule
{\bf Total} & & & & & 789824 \\
\bottomrule

\end{tabular}
\egroup
%\vspace{0.2cm}
\caption{{\bf Transformer architecture.} The network receives as input $X_o$ and outputs $Z$. All MLP's employ a ReLU nonlinearity. $D$ is the number of feature dimensions; $H$ is the number of attention heads. In our model $D$=256, $H$=4 and $k$=4.}
\label{table:transformer}
\end{table*}

\clearpage
\newpage

\section{Additional Motion Prediction Results}

% jngiam comment: I don't think validation set numbers are fair to compare against, our numbers seem too low and most papers don't compare them (unless their test numbers don't look as good).

% At Step 1.14M
% https://chauffeur-mldash.corp.google.com/experiments/8718669757441326514#scalars&tagFilter=minADE&runSelectionState=eyJ0cmFpbl90cmFpbiI6ZmFsc2UsImRlY29kZV9kZXZicHZpeiI6ZmFsc2V9&_smoothingWeight=0

% \begin{table}[t]
% \centering
% \footnotesize
% \begin{tabular}{c|ccc}
%     \toprule
%     Validation & minADE$\,\downarrow$ & minFDE$\,\downarrow$ & MR$\,\downarrow$ \\
%     \midrule
%      mmTransformer \cite{liu2021mmformer} & \update{0.713} & \update{1.153} & \update{0.106} \\
%      TNT \cite{zhao2020tnt} & \update{0.73} & \update{1.29} & \update{0.09} \\
%      LaneGCN \cite{liang2020laneGCN} & \update{0.71} & \update{1.08} & - \\
%      TPCN \cite{ye2021tpcn} & 0.73 & 1.15 & 0.11\\
%      ours (marginal) & 0.645 & 1.09 & 0.10 \\
%      \bottomrule
% \end{tabular}
% \vspace{0.2cm}
% \label{table:argoverse-validation-results}
% \caption{\textbf{Predictive performance on Argoverse motion prediction.} Results reported on {\it validation} split for vehicles \cite{chang2019argoverse}. Reported are the minADE, minFDE for $k=6$ predictions \cite{alahi2016social,pellegrini2009you}, and the Miss Rate (MR) \cite{chang2019argoverse} within 2 meters of the target. Note that marginal results are only shown because the evaluation only predicts one agent.}
% \end{table}

% t = 3
\begin{table}[h]
\centering
\footnotesize
\bgroup
\def\arraystretch{1.05}%  1 is the default.
\resizebox{\linewidth}{!}{
\begin{tabular}{cc|ccc|ccc|ccc|ccc}
    \toprule
    \multicolumn{2}{c|}{\multirow{2}{*}{{\bf Motion Prediction}}} & \multicolumn{3}{c|}{minADE$\,\downarrow$} & \multicolumn{3}{c|}{minFDE$\,\downarrow$} & \multicolumn{3}{c|}{MR$\,\downarrow$} & \multicolumn{3}{c}{mAP$\,\uparrow$}\\
    & & veh & ped & cyc & veh & ped & cyc & veh & ped & cyc & veh & ped & cyc\\ 
    \midrule
    {\it valid} & & & & & & & & & & &\\
     & baseline \cite{ettinger2021large} & 0.39 & 0.19 & 0.41 & 0.65 & 0.36 & 0.73 & 0.14 & 0.07 & 0.25 & 0.33 & \textbf{0.33} & 0.27\\
     % At 857.6K steps
     % https://tensorboard.corp.google.com/experiment/mldash:chauffeur:6508684537221286330/#scalars&tagFilter=marginal_metrics%2F.*_at_8s%2FminADE&_smoothingWeight=0&runSelectionState=eyJkZWNvZGVfZGV2YnB2aXoiOmZhbHNlLCJldmFsX2RldmJwL0RldkJQSW50ZXJhY3RpdmUiOmZhbHNlfQ%3D%3D
     & ours (marginal) & {\bf 0.33} & \textbf{0.20} & \textbf{0.39} & \textbf{0.57} & \textbf{0.33} & \textbf{0.67} & \textbf{0.11} & \textbf{0.07} & \textbf{0.21} & \textbf{0.38} & \textbf{0.33} & \textbf{0.28}\\
     % At 857.6K steps
     % https://tensorboard.corp.google.com/experiment/mldash:chauffeur:8119650759533646213/
     & ours (joint-as-marginal) & 0.42 & 0.28 & 0.50 & 0.78 & 0.51 & 0.94 & 0.19 & 0.21 & 0.32 & 0.34 & 0.25 & 0.23\\
     % At 371.2K steps
     % https://tensorboard.corp.google.com/experiment/mldash:chauffeur:5237978701955273581/#scalars&regexInput=Dev&tagFilter=joint_metrics%2FVehicle_at_8s&_smoothingWeight=0
     & ours (multi-task joint-as-marginal) & 0.43 & 0.51 & 0.51 & 0.80 & 0.52 & 0.93 & 0.20 & 0.21 & 0.32 & 0.33 & 0.24 & 0.23\\
     \midrule
    {\it test} & & & & & & & & & & & \\
     & baseline \cite{ettinger2021large} & 0.39 & 0.20 & 0.41 & 0.65 & 0.36 & 0.74 & 0.14 & 0.07 & 0.25 & 0.34 & \textbf{0.32} & 0.24\\
     % At 857.6K steps
     & ours (marginal) & \textbf{0.32} & \textbf{0.20} & \textbf{0.38} & \textbf{0.56} & \textbf{0.33} & \textbf{0.67} & \textbf{0.11} & \textbf{0.07} & \textbf{0.21} & \textbf{0.38} & \textbf{0.32} & \textbf{0.28} \\
     & ours (joint-as-marginal) & 0.42 & 0.28 & 0.49 & 0.78 & 0.53 & 0.92 & 0.19 & 0.21 & 0.32 & 0.33 & 0.26 & 0.24 \\
     & ours (multi-task joint-as-marginal) & 0.44 & 0.29 & 0.50 & 0.80 & 0.53 & 0.93 & 0.20 & 0.22 & 0.32 & 0.32 & 0.24 & 0.24\\
     \bottomrule
\end{tabular}}
\egroup
\vspace{0.2cm}
\caption{\textbf{Marginal predictive performance on Waymo Open Motion Dataset motion prediction for $t = 3$ seconds}. Please see Table \ref{table:wmod-marginal-results} for details. \label{table:wmod-results-marginal-t3}}
\end{table}

\begin{table}[h]
\bgroup
\def\arraystretch{1.05}%  1 is the default.
\resizebox{\linewidth}{!}{
\begin{tabular}{cc|ccc|ccc|ccc|ccc}
    \toprule
    \multicolumn{2}{c|}{\multirow{2}{*}{{\bf Interaction Prediction}}} & \multicolumn{3}{c|}{minSADE$\,\downarrow$} & \multicolumn{3}{c|}{minSFDE$\,\downarrow$} & \multicolumn{3}{c|}{SMR$\,\downarrow$} & \multicolumn{3}{c}{mAP$\,\uparrow$}\\
    & & veh & ped & cyc & veh & ped & cyc & veh & ped & cyc & veh & ped & cyc\\ 
    \midrule
    {\it valid} & & & & & & & & & & &\\
     & baseline \cite{ettinger2021large} & 0.58 & 0.43 & 0.60 & 1.13 & 0.86 & 1.20 & 0.45 & 0.47 & 0.61 & 0.15 & 0.13 & 0.06\\
     % At 902.4K steps
     & ours (marginal-as-joint) & 0.45 & 0.37 & 0.52 & 0.91 & 0.75 & 1.04 & 0.35 & 0.42 & 0.52 & 0.20 & 0.12 & 0.09\\
     % At 857.6K steps
     % https://tensorboard.corp.google.com/experiment/mldash:chauffeur:8119650759533646213/#scalars     
     & ours (joint) & 0.41 & \textbf{0.34} & \textbf{0.47} & 0.81 & \textbf{0.65} & 0.92 & \textbf{0.29} & \textbf{0.38} & \textbf{0.49} & 0.26 & 0.14 & 0.11 \\
     % At 371.2K steps
     % https://tensorboard.corp.google.com/experiment/mldash:chauffeur:5237978701955273581/#scalars&regexInput=Dev&tagFilter=joint_metrics%2FVehicle_at_8s&_smoothingWeight=0
     & ours (multi-task joint) & \textbf{0.40} & \textbf{0.34} & \textbf{0.47} & \textbf{0.80} & \textbf{0.65} & \textbf{0.91} & \textbf{0.28} & \textbf{0.38} & \textbf{0.49} & \textbf{0.27} & \textbf{0.17} & \textbf{0.13}\\     
     \midrule
    {\it test} & & & & & & & & & & & \\
     & baseline \cite{ettinger2021large} & 0.58 & 0.42 & 0.61 & 1.14 & 0.85 & 1.21 & 0.45 & 0.47 & 0.61 & 0.16 & 0.11 & 0.05 \\
     % At 857.6K steps
     & ours (marginal-as-joint) & 0.45 & 0.36 & 0.53 & 0.93 & 0.74 & 1.06 & 0.36 & 0.40 & 0.55  & 0.18 & 0.11 & 0.07 \\
     & ours (joint) & \textbf{0.41} & \textbf{0.33} & \textbf{0.48} & 0.82 & \textbf{0.64} & \textbf{0.94} & \textbf{0.29} & \textbf{0.36} & \textbf{0.50} & 0.18 & 0.12 & 0.07\\
     & ours (multi-task joint) & \textbf{0.41} & 0.34 & \textbf{0.48} & \textbf{0.81} & 0.65 & \textbf{0.94} & \textbf{0.29} & 0.37 & 0.51   & \textbf{0.26} & \textbf{0.15} & \textbf{0.10}\\
     \bottomrule
\end{tabular}}
\egroup
\caption{\textbf{Joint predictive performance on Waymo Open Motion Dataset motion prediction for $t = 3$ seconds}. Please see Table \ref{table:wmod-joint-results} for details. \label{table:wmod-results-joint-t3}}
\end{table}

% t = 5
\begin{table}[h]
\centering
\footnotesize
\bgroup
\def\arraystretch{1.05}%  1 is the default.
\resizebox{\linewidth}{!}{
\begin{tabular}{cc|ccc|ccc|ccc|ccc}
    \toprule
    \multicolumn{2}{c|}{\multirow{2}{*}{{\bf Motion Prediction}}} & \multicolumn{3}{c|}{minADE$\,\downarrow$} & \multicolumn{3}{c|}{minFDE$\,\downarrow$} & \multicolumn{3}{c|}{MR$\,\downarrow$} & \multicolumn{3}{c}{mAP$\,\uparrow$}\\
    & & veh & ped & cyc & veh & ped & cyc & veh & ped & cyc & veh & ped & cyc\\ 
    \midrule
    {\it valid} & & & & & & & & & & &\\
     & baseline \cite{ettinger2021large} & 0.74 & 0.37 & 0.75 & 1.36 & 0.73 & 1.43 & 0.17 & 0.10 & 0.25 & 0.29 & \textbf{0.27} & \textbf{0.26}\\
     % At 857.6K steps
     % https://tensorboard.corp.google.com/experiment/mldash:chauffeur:6508684537221286330/#scalars&tagFilter=marginal_metrics%2F.*_at_8s%2FminADE&_smoothingWeight=0&runSelectionState=eyJkZWNvZGVfZGV2YnB2aXoiOmZhbHNlLCJldmFsX2RldmJwL0RldkJQSW50ZXJhY3RpdmUiOmZhbHNlfQ%3D%3D
     & ours (marginal) & {\bf 0.65} & \textbf{0.35} & \textbf{0.69} & \textbf{1.23} & \textbf{0.67} & \textbf{1.30} & \textbf{0.15} & \textbf{0.10} & \textbf{0.22} & \textbf{0.33} & 0.26 & 0.25\\
     % At 857.6K steps
     % https://tensorboard.corp.google.com/experiment/mldash:chauffeur:8119650759533646213/
     & ours (joint-as-marginal) & 0.83 & 0.51 & 0.93 & 1.68 & 1.08 & 1.91 & 0.23 & 0.26 & 0.33 & 0.26 & 0.19 & 0.18 \\
     % At 371.2K steps
     % https://tensorboard.corp.google.com/experiment/mldash:chauffeur:5237978701955273581/#scalars&regexInput=Dev&tagFilter=joint_metrics%2FVehicle_at_8s&_smoothingWeight=0
     & ours (multi-task joint-as-marginal) & 0.84 & 0.52 & 0.92 & 1.72 & 1.08 & 1.87 & 0.24 & 0.26 & 0.33 & 0.26 & 0.18 & 0.18 \\
     \midrule
    {\it test} & & & & & & & & & \\
     & baseline \cite{ettinger2021large} & 0.74 & 0.37 & 0.76 & 1.35 & 0.73 & 1.43 & 0.17 & 0.10 & 0.25 & 0.29 & \textbf{0.26} & \textbf{0.23}\\
     % At 857.6K steps
     & ours (marginal) & \textbf{0.63} & \textbf{0.35} & \textbf{0.68} & \textbf{1.20} & \textbf{0.67} & \textbf{1.31} & \textbf{0.11} & \textbf{0.10} & \textbf{0.22} & \textbf{0.33} & \textbf{0.26} & \textbf{0.23} \\
     & ours (joint-as-marginal) & 0.83 & 0.52 & 0.91 & 1.68 & 1.10 & 1.88 & 0.23 & 0.26 & 0.33 & 0.26 & 0.19 & 0.22\\
     & ours (multi-task joint-as-marginal) & 0.85 & 0.53 & 0.93 & 1.72 & 1.10 & 1.91 & 0.24 & 0.27 & 0.34 & 0.26 & 0.19 & 0.21\\
     \bottomrule
\end{tabular}}
\egroup
\caption{\textbf{Marginal predictive performance on Waymo Open Motion Dataset motion prediction for $t = 5$ seconds}. Please see Table \ref{table:wmod-marginal-results} for details. \label{table:wmod-results-marginal-t5}}
\end{table}

\begin{table}[h]
\bgroup
\def\arraystretch{1.05}%  1 is the default.
\resizebox{\linewidth}{!}{
\begin{tabular}{cc|ccc|ccc|ccc|ccc}
    \toprule
    \multicolumn{2}{c|}{\multirow{2}{*}{{\bf Interaction Prediction}}} & \multicolumn{3}{c|}{minSADE$\,\downarrow$} & \multicolumn{3}{c|}{minSFDE$\,\downarrow$} & \multicolumn{3}{c|}{SMR$\,\downarrow$} & \multicolumn{3}{c}{mAP$\,\uparrow$}\\
    & & veh & ped & cyc & veh & ped & cyc & veh & ped & cyc & veh & ped & cyc\\ 
    \midrule
    {\it valid} & & & & & & & & & & &\\
     & baseline \cite{ettinger2021large} & 1.19 & 0.90 & 1.25 & 2.64 & 1.98 & 2.82 & 0.55 & 0.57 & 0.70 & 0.13 & 0.09 & 0.04\\
     % At 902.4K steps
     & ours (marginal-as-joint) & 0.96 & 0.78 & 1.08 & 2.15 & 1.75 & 2.46 & 0.44 & 0.51 & 0.60 & 0.13 & 0.09 & 0.06 \\
     % At 857.6K steps
     % https://tensorboard.corp.google.com/experiment/mldash:chauffeur:8119650759533646213/#scalars     
     & ours (joint) & 0.84 & 0.68 & 0.95 & 1.81 & 1.45 & 2.09 & \textbf{0.38} & \textbf{0.48} & \textbf{0.59} & 0.18 & 0.08 & \textbf{0.07} \\
     % At 371.2K steps
     % https://tensorboard.corp.google.com/experiment/mldash:chauffeur:5237978701955273581/#scalars&regexInput=Dev&tagFilter=joint_metrics%2FVehicle_at_8s&_smoothingWeight=0
     & ours (multi-task joint) & \textbf{0.83} & \textbf{0.68} & \textbf{0.95} & 1.79 & 1.45 & \textbf{2.08} & \textbf{0.38} & 0.48 & \textbf{0.59} & \textbf{0.19} & \textbf{0.10} & 0.06\\     
     \midrule
    {\it test} & & & & & & & & & \\
     & baseline \cite{ettinger2021large} & 1.21 & 0.89 & 1.26 & 2.70 & 1.96 & 2.80 & 0.56 & 0.59 & 0.69 & 0.13 & 0.07 & 0.03 \\
     % At 857.6K steps
     & ours (marginal-as-joint) & 0.96 & 0.77 & 1.07  & 2.19 & 1.74 & 2.45 & 0.46 & 0.51 & 0.62 & 0.13 & 0.09 & 0.06 \\
     & ours (joint) & 0.85 & \textbf{0.67} & \textbf{0.96} & 1.85 & \textbf{1.44} & 2.10 & \textbf{0.39} & 0.50 & \textbf{0.58} & 0.17 & \textbf{0.10} & 0.07\\
     & ours (multi-task joint) & \textbf{0.84} & 0.68 & 0.97 & \textbf{1.82} & 1.46 & \textbf{2.09} & \textbf{0.39} & \textbf{0.49} & 0.59 & \textbf{0.20} & 0.09 & \textbf{0.08}\\
     \bottomrule
\end{tabular}}
\egroup
\caption{\textbf{Joint predictive performance on Waymo Open Motion Dataset motion prediction for $t = 5$ seconds}. Please see Table \ref{table:wmod-joint-results} for details. \label{table:wmod-results-joint-t5}}
\end{table}

% TODO(bencaine, anyone): Make this one table again?
\begin{table}[h]
\centering
\footnotesize
\bgroup
\def\arraystretch{1.05}%  1 is the default.
\resizebox{\linewidth}{!}{
\begin{tabular}{cc|ccc|ccc|ccc|ccc}
    \toprule
    \multicolumn{2}{c|}{\multirow{2}{*}{{\bf Motion Prediction}}} & \multicolumn{3}{c|}{minADE$\,\downarrow$} & \multicolumn{3}{c|}{minFDE$\,\downarrow$} & \multicolumn{3}{c|}{MR$\,\downarrow$} & \multicolumn{3}{c}{mAP$\,\uparrow$} \\
    & & veh & ped & cyc & veh & ped & cyc & veh & ped & cyc & veh & ped & cyc\\ 
    \midrule
    {\it valid} & & & & & & & & & & & \\
     & LSTM baseline \cite{ettinger2021large} & 1.34 & 0.63 & 1.26 & 2.85 & 1.35 & 2.68 & 0.25 & 0.13 & 0.29 & 0.23 & 0.23 & 0.20 \\
     % From leaderboard submission
     & ours (marginal) & {\bf 1.17} & \textbf{0.59} & \textbf{1.15} & \textbf{2.51} & \textbf{1.26} & \textbf{2.44} & \textbf{0.20} & \textbf{0.12} & \textbf{0.24} & 0.26 & 0.27 & 0.20 \\
    %  & ours (marginal + NMS) & 1.23 & 0.75 & 1.25 & 2.64 & 1.68 & 2.68 & 0.20 & 0.24 & 0.27 & \textbf{0.33} & \textbf{0.26} & \textbf{0.26} \\
     % From leaderboard submission
     & ours (joint-as-marginal) & 1.53 & 0.90 & 1.63 & 3.48 & 2.09 & 3.77 & 0.28 & 0.30 & 0.37 & 0.20 & 0.16 & 0.13\\
     % At 371.2K steps
     % https://tensorboard.corp.google.com/experiment/mldash:chauffeur:5237978701955273581/#scalars&regexInput=Dev&tagFilter=joint_metrics%2FVehicle_at_8s&_smoothingWeight=0
     & ours (multi-task joint-as-marginal) & 1.56 & 0.89 & 1.60 & 3.56 & 2.06 & 3.68 & 0.29 & 0.29 & 0.37 & 0.19 & 0.15 & 0.15\\
     \midrule
    {\it test} & & & & & & & & & & & & \\
     & LSTM baseline \cite{ettinger2021large} & 1.34 & 0.64 & 1.29 & 2.83 & 1.35 & 2.68 & 0.24 & 0.13 & 0.29 & 0.24 & 0.22 & 0.19\\
     & ours (marginal) & \textbf{1.17} & \textbf{0.60} & \textbf{1.17} & \textbf{2.48} & \textbf{1.25} & \textbf{2.43} & \textbf{0.19} & \textbf{0.12} & \textbf{0.22} & 0.27 & 0.23 & 0.20\\
    %  & ours (marginal + NMS) & 1.22 & 0.76 & 1.27 & 2.61 & 1.68 & 2.69 & 0.20 & 0.23 & 0.27 & \textbf{0.33} & \textbf{0.27} & \textbf{0.25}\\
     & ours (joint-as-marginal) & 1.52 & 0.91 & 1.61 & 3.43 & 2.09 & 3.68 & 0.28 & 0.30 & 0.37 & 0.19 & 0.16 & 0.19\\
     & ours (multi-task joint-as-marginal) & 1.55 & 0.91 & 1.64 & 3.50 & 2.08 & 3.75 & 0.28 & 0.29 & 0.38 & 0.19 & 0.16 & 0.19\\
     \bottomrule
\end{tabular}}
\egroup
\caption{\textbf{Marginal predictive performance on Waymo Open Motion Dataset motion prediction for $t = 8$ seconds}. Please see Table \ref{table:wmod-marginal-results} for details. We additionally include "joint-as-marginal" results for our standard MP masked joint model, and our multi-task joint model. These are joint models evaluated as if they were marginal models (with no changes in the outputs). \label{table:wmod-results-marginal-t8}}
\vspace{-0.1cm}
\end{table}

\begin{table}[h]
\centering
\footnotesize
\bgroup
\def\arraystretch{1.05}%  1 is the default.
\resizebox{\linewidth}{!}{
\begin{tabular}{cc|ccc|ccc|ccc|ccc}
    \toprule
    \multicolumn{2}{c|}{\multirow{2}{*}{{\bf Interaction Prediction}}} & \multicolumn{3}{c|}{minSADE$\,\downarrow$} & \multicolumn{3}{c|}{minSFDE$\,\downarrow$} & \multicolumn{3}{c|}{SMR$\,\downarrow$} & \multicolumn{3}{c}{mAP$\,\uparrow$}\\
    & & veh & ped & cyc & veh & ped & cyc & veh & ped & cyc & veh & ped & cyc\\ 
    \midrule
    {\it valid} & & & & & & & & & & &\\
     & LSTM baseline \cite{ettinger2021large} & 2.42 & 2.73 & 3.16 & 6.07 & 4.20 & 6.46 & 0.66 & 1.00 & 0.83 & 0.07 & \textbf{0.06} & 0.02 \\
     % From leaderboard submission
     & ours (marginal-as-joint) & 2.04 & 1.62 & 2.28 & 4.94 & 3.81 & 5.67 & 0.54 & 0.63 & 0.72 & \textbf{0.11} & 0.05 & 0.03 \\
     % From leaderboard submission
     & ours (joint, mp-only) & \textbf{1.72} & \textbf{1.38} & 1.96 & \textbf{3.98} & \textbf{3.11} & 4.75 & \textbf{0.49} & \textbf{0.60} & 0.73 & \textbf{0.11} & 0.05 & 0.03  \\
     % At 371.2K steps
     % https://tensorboard.corp.google.com/experiment/mldash:chauffeur:5237978701955273581/#scalars&regexInput=Dev&tagFilter=joint_metrics%2FVehicle_at_8s&_smoothingWeight=0
     & ours (joint, multi-task) & \textbf{1.72} & 1.39 & \textbf{1.94} & 3.99 & 3.15 & \textbf{4.69} & \textbf{0.49} & 0.62 & \textbf{0.71} & \textbf{0.11} & \textbf{0.06} & \textbf{0.04} \\     
     \midrule
    {\it test} & & & & & & & & & & & \\
     & LSTM baseline \cite{ettinger2021large} & 2.46 & 2.47 & 2.96 & 6.22 & 4.30 & 6.26 & 0.67 & 0.89 & 0.89 & 0.06 & 0.03 & 0.03\\
     % At 857.6K steps
     & ours (marginal-as-joint) & 2.08 & 1.62 & 2.24 & 5.04 & 3.87 & 5.41 & 0.55 & 0.64 & 0.73 & 0.08 & \textbf{0.05} & 0.03 \\
     & ours (joint, mp-only) & 1.76 & \textbf{1.38} & \textbf{1.95} & 4.08 & \textbf{3.19} & \textbf{4.65} & \textbf{0.50} & \textbf{0.62} & \textbf{0.70} & 0.10 & \textbf{0.05} & \textbf{0.04}\\
     & ours (joint, multi-task) & \textbf{1.74} & 1.41 & \textbf{1.95} & \textbf{4.06} & 3.26 & 4.68 & \textbf{0.50} & 0.64 & 0.71 & \textbf{0.13} & 0.04 & 0.03\\
     \bottomrule
\end{tabular}}
\egroup
%\vspace{0.1cm}
\caption{\textbf{Joint predictive performance on Waymo Open Motion Dataset motion prediction for $t = 8$ seconds}. These results are the same as Table \ref{table:wmod-joint-results} but is included here for completeness with the other Appendix tables.  \label{table:wmod-results-joint-t8}}
\vspace{-0.3cm}
\end{table}

\clearpage

\subsection{Understanding the trade-offs between the joint and marginal models}\label{tradeoffs}

% NOTE(bencaine): I think these graphics are really confusing because we show marginal minADE for both the
% marginal and joint model. 
\begin{figure*}[h]
\centering
% \includegraphics[width=0.3\textwidth]{figures/joint_marginal_minade.png} %/Overlap-Scatter.jpg} 
% \hspace{0.1cm}
%\includegraphics[width=0.24\textwidth]{figures/joint_marginal_overlap.png} %Overlap-Difference.pdf}
%\includegraphics[width=0.24\textwidth]{figures/slicing/original_minADE_average_sdc_speed_slicing_cut.png}
\includegraphics[width=0.30\textwidth]{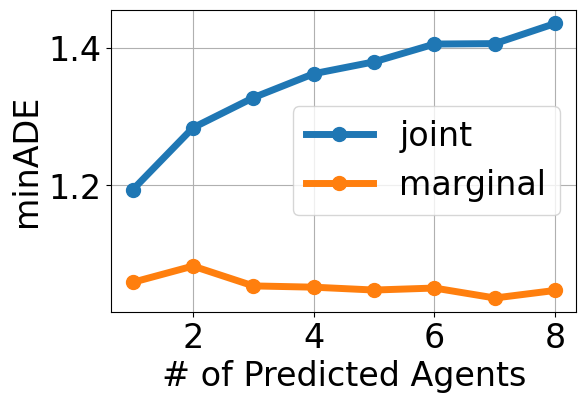}
\hspace{0.2cm}
\includegraphics[width=0.30\textwidth]{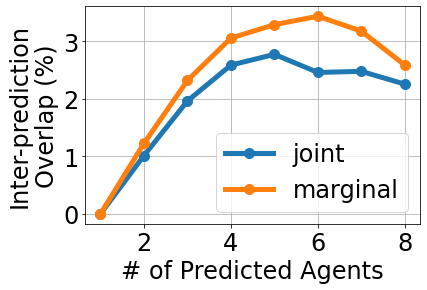}
\vspace{-0.3cm}
\caption{\textbf{Quantitative comparison of the marginal and joint prediction models.} Left: The \textit{marginal} minADE broken down as a function of number of predicted agents. When there are more agents, producing internally consistent predictions is more challenging, and hence the joint model performs slightly worse. Right: Overlap rate between pairs of agent predictions. The joint model produces internally consistent predictions with lower inter-prediction overlap rates. \label{figure:joint-marginal-comparison}}
% \caption{\textbf{Quantitative comparison of the marginal and joint prediction models.} Left: Cumulative distribution of the minADE for the marginal and joint models evaluated on the motion prediction task. The mean minADE for the marginal model is lower than the joint model (marginal $=1.05m$ vs joint $=1.35m$). Center: The minADE broken down as a function of number of predicted agents. When there are more agents, producing internally consistent predictions is more challenging, and hence the joint model performs slightly worse. Right: Overlap rate between pairs of agent predictions. The joint model produces internally consistent predictions with lower inter-prediction overlap rates. \label{figure:joint-marginal-comparison}}
%Comparison of the overlap metric between the marginal and joint prediction model. 
%Plot shows the number of predicted trajectories with a non-zero overlap with the bounding box of any other agent within the scenario as a fraction of the number of predicted agents within the scenario. Each dot represents the number of overlaps that occur on a single scenario for the joint and marginal models, respectively. The line denotes line of equality while the green + marks the mean overlap fraction for joint and marginal models at \todo{0.010462 and 0.021274} respectively. Right:  \label{figure:joint-marginal-comparison}}
\end{figure*}

While the joint mp-only model performs better on the interactive prediction task than the marginal model \textit{when evaluated with joint metrics}, it performs worse on the motion prediction task \textit{when evaluated with marginal metrics} (see Table \ref{table:wmod-results-marginal-t8} "ours (joint)"). This is because the joint model is trained to produce predictions that are internally consistent between agents, while the marginal model does not, which is a strictly harder task. In this section, we examine this quality difference and the internal consistency of the predictions from both models.

In the \wmod dataset, each example has a different number of agents to be predicted. By slicing the {\it marginal} minADE results based on the number of agents to be predicted (Figure \ref{figure:joint-marginal-comparison}, left), we find that the joint model performs worse as there are more predicted agents, while the marginal model performs the same. This is expected: when there are more agents, the joint model has a more difficult task since it needs to produce internally consistent predictions. One may expect a joint model to need an exponential number of trajectory combinations to perform competitively with the marginal model, but we are encouraged to find that this is not actually the case. We believe this is due to the fact that when many agents are interacting, the number of realistic scenarios is actually heavily constrained by interactions - most combinations of marginal predictions don't actually make sense together.

However, when more agents are to be predicted, the possibility of interactions are higher and we would expect that the joint model is able to capture these interactions through internally consistent predictions. We measure the models' internal consistency by selecting the best trajectory and measuring the {\it inter-prediction overlap rate} (details below). We find that the joint model has a consistently lower inter-prediction overlap rate, showing that it is able to capture the interactions between agents. The ability to model interactions enables the model to be suitable for conditional motion prediction, and goal conditioned prediction tasks, which we discuss in Section \ref{multitask}.

{\bf Measuring agent inter-prediction overlap.} We measure \emph{predicted} agent overlap on the best set of trajectory produced by the model. For the joint model, this corresponds to the joint prediction that has the higher probability score. For the marginal model, we take the top scoring trajectory for each agent to obtain the best set. For every predicted agent in the trajectory, we determine if it has an overlap with any \textit{other predicted} agent by comparing the rotated upright 3D bounding boxes using our predicted xyz and heading. The inter-prediction overlap rate is the number of predicted agents that are involved in some overlap, divided by the total number of predicted agents. We count two agents as overlapping if the intersection over the union (IoU) exceeds 0.01.

\newpage
\clearpage
\section{Slicing results \label{section:slicing-results}}
\begin{figure}[ht!]
    \centering
    \includegraphics[width=0.48\textwidth]{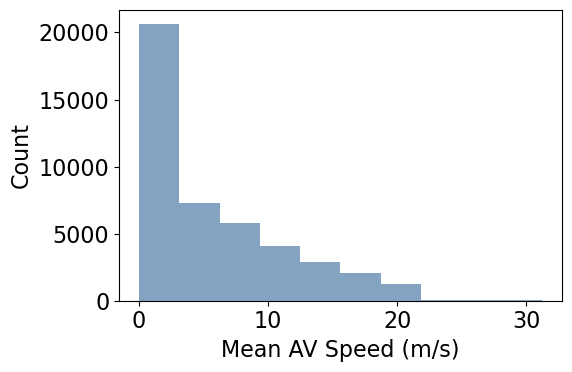} \includegraphics[width=0.48\textwidth]{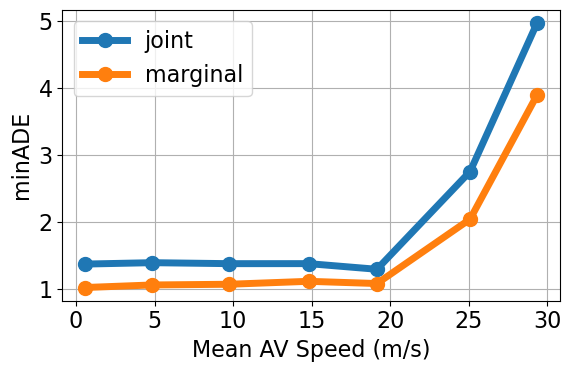} \\
        \includegraphics[width=0.48\textwidth]{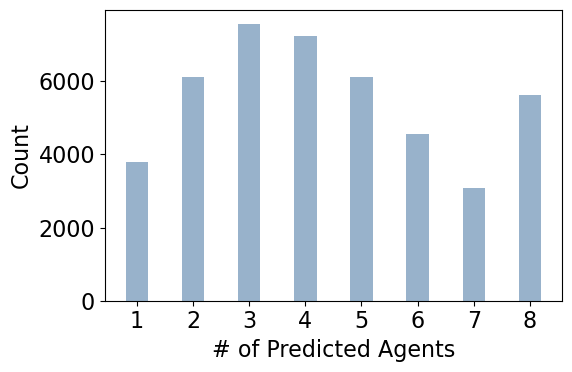} 
        \includegraphics[width=0.48\textwidth]{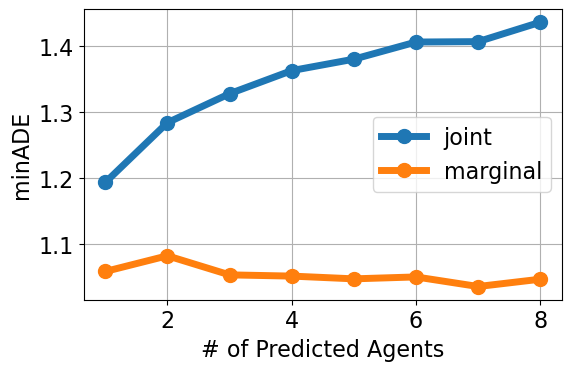} \\
            \includegraphics[width=0.48\textwidth]{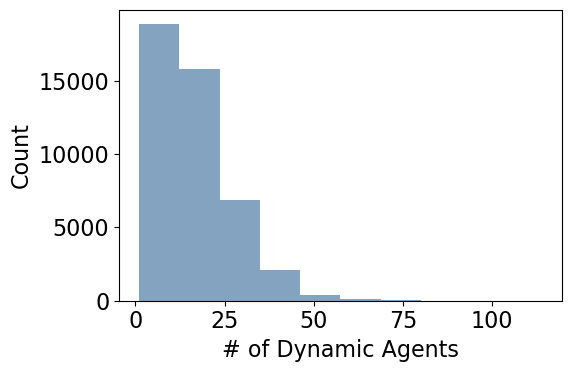}
            \includegraphics[width=0.48\textwidth]{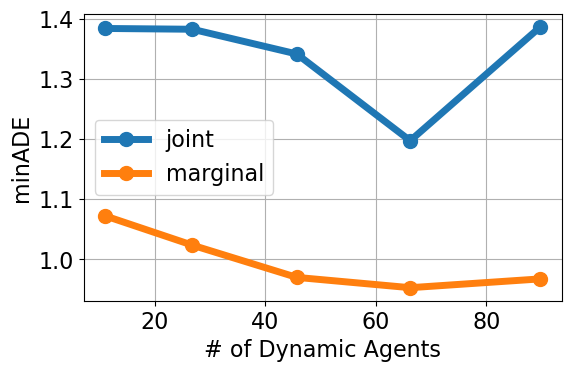} \\
    \caption{We show marginal minADE on a per scene basis broken down by different scene level statistics. Interestingly, we show that both the marginal and joint models become out of distribution above 20 m/s, where there is minimal training data.}
    \label{fig:my_label}
\end{figure}
% \begin{figure}[]
%    \centering
%    \includegraphics[width=0.48\textwidth]{figures/Overlap-Scatter.jpg} 
%    \includegraphics[width=0.48\textwidth]{figures/Overlap-Difference.png} \\
%    \caption{Relative performance of the marginal and joint models across individual scenes.}
%    \label{fig:individual_scenes}
%\end{figure}

\begin{figure*}[h]
\centering
\includegraphics[width=0.45\textwidth]{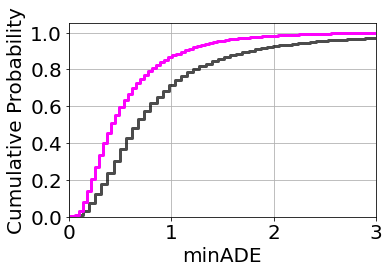}
\includegraphics[width=0.45\textwidth]{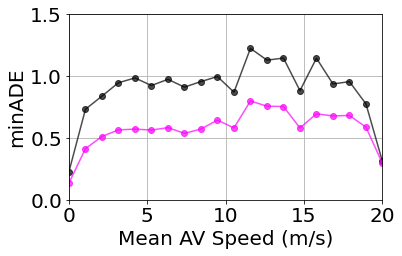}
\caption{\textbf{Analysis of the multi-task model for goal-conditioned motion prediction.} Left: Cumulative probability of the AV minADE in goal conditioned prediction (blue) and motion prediction (black) masking strategies. 
Right: AV minADE for goal conditioned prediction (blue) and motion prediction (black) as a function of AV speed (averaged over the ground truth trajectory).
\label{fig:planning-quantitative}}
\end{figure*}

\newpage
\clearpage
\section{Loss Implementation \label{section:loss-code}}

The Scene Transformer model is both agent permutation equivariant and scene-centric. These properties allow us to easily switch between a marginal vs joint loss formulation (Figure \ref{fig:loss-code}). In the marginal formulation, we reduce the loss of the best future for every agent separately; while in the joint formulation, we reduce the loss of the best future for \textit{all} agents jointly. In practice, this determines when the reduce\_min operation is performed.

Our final loss composes the regression loss on the trajectory and the classification loss of the best trajectory. A weighted linear combination of the loss terms is used to combine these two losses together. The classification loss weight was set to be 0.1, while the regression losses have weight 1.0. The weights were determined using the hold-out validation set.

\begin{figure}[h!]
\centering
\begin{minted}[fontsize=\footnotesize]{python}

# agent_predictions and agent_gt are 
# [F, A, T, 7] Tensors with [x, y, z], uncertainty terms, and yaw.
 
# Get the KL Divergence of the predictions vs ground truth.
# Use LaplaceKL method from (Meyer & Thakurdesai, 2020)
loss = LaplaceKL(agent_predictions, agent_gt)
 
# Now reduce across all timesteps and values to produce
# a tensor of shape [F, A]
loss = tf.reduce_sum(loss, axis=[2, 3])

# The marginal loss, we only apply the loss to the best trajectory
# per agent (so min across the future dimension).
marginal_loss = tf.reduce_min(loss, axis=0)
 
# Then sum over the agent dimension
marginal_loss = tf.reduce_sum(marginal_loss)

# The joint loss, we sum over all agents to get a loss value
# per future.
joint_loss = tf.reduce_sum(loss, axis=1)
 
# Then only apply the loss to the best future prediction
joint_loss = tf.reduce_min(joint_loss)

\end{minted}
\caption{Pseudo-code in TensorFlow \cite{tensorflow2015-whitepaper} demonstrating the joint versus marginal loss formulation.}
\label{fig:loss-code}
\end{figure}

\end{document}